%% file: egpaper_for_review.tex
\documentclass[10pt,twocolumn,letterpaper]{article}

\usepackage{iccv}
\usepackage{times}
\usepackage{epsfig}
\usepackage{graphicx}
\usepackage{amsmath}
\usepackage{amssymb}
\usepackage{subfig}

\usepackage[pagebackref=true,breaklinks=true,letterpaper=true,colorlinks,bookmarks=false]{hyperref}

 \iccvfinalcopy 

\def\iccvPaperID{38} 

\ificcvfinal\fi

\newcommand{\convnets}{convolutional neural networks\xspace}

\begin{document}

\title{Contextual Interference Reduction by Selective Fine-Tuning of Neural Networks}

\author{Mahdi Biparva, John Tsotsos\\
	Department of Electrical Engineering and Computer Science\\
	York University\\
	Toronto, Canada\\
	{\tt\small \{mhdbprv,tsotsos\}@cse.yorku.ca}}

\maketitle
\ificcvfinal\thispagestyle{empty}\fi

\begin{abstract}
	\input{00-abstract.tex}
\end{abstract}

\section{Introduction}
\label{sec:int}
\input{01-introduction.tex}

\section{Selective Attention for Network Fine-Tuning}
\label{sec:mod}
\input{03-model.tex}

\section{Experimental Results}
\label{sec:exp}

\input{04-experiments.tex}

\section{Conclusion}
\label{sec:con}
\input{05-conclusion.tex}

{\small
\bibliographystyle{ieee}
\bibliography{thesis_phd}
}

\end{document}

%% file: 00-abstract.tex
Feature disentanglement of the foreground target objects and the background surrounding context has not been yet fully accomplished. The lack of network interpretability prevents advancing for feature disentanglement and better generalization robustness. We study the role of the context on interfering with a disentangled foreground target object representation in this work. We hypothesize that the representation of the surrounding context is heavily tied with the foreground object due to the dense hierarchical parametrization of convolutional networks with under-constrained learning algorithms. Working on a framework that benefits from the bottom-up and top-down processing paradigms, we investigate a systematic approach to shift learned representations in feedforward networks from the emphasis on the irrelevant context to the foreground objects. The top-down processing provides importance maps as the means of the network internal self-interpretation that will guide the learning algorithm to focus on the relevant foreground regions towards achieving a more robust representations. We define an experimental evaluation setup with the role of context emphasized using the MNIST dataset. The experimental results reveal not only that the label prediction accuracy is improved but also a higher degree of robustness to the background perturbation using various noise generation methods is obtained.

%% file: 01-introduction.tex

The issue of the contextual interference with the foreground target objects is one of the shortcomings of the hierarchical feature representations such as convolutional neural networks. 
the foreground and background representations are inevitably mixed up and visual confusion is eminent due to the dense hierarchical parametrization of convolutional networks and the under-constrained utilization of convolution and sub-sampling layers in the feedforward manner. 
Feedforward neural networks trained for object classification have shown successful application of localization through Top-Down mechanisms. Despite the success of localizing objects in the cluttered natural images using such feedforward networks, the context has still significant role in the final label prediction \cite{dodge2017can,biparva2017stnet,rosenfeld2018elephant}. Additionally, research studies have revealed evidence on widespread visual confusion on \convnets \cite{nguyen2015deep,dodge2017can,akhtar2018threat,rosenfeld2018elephant,kurakin2016adversarial}.
A systematic approach to shift learned neural representations from the emphasis on the contextual regions to the foreground target objects can help achieve a higher degree of representation disentanglement.
We propose a selective fine-tuning approach for neural networks using a unified bottom-up and top-down framework. 
A gating mechanism of hidden activities imposed by Top-Down selection mechanisms is defined in the iterative feedforward pass. 
An attention-augmented loss function is introduced during which the network parameters are fine-tuned for a number of iterations. The fine-tuning using the iterative pass helps the network to reduce the reliance on the contextual representation throughout the visual hierarchy. 
Therefore, the label prediction relies more on the target object representation and consequently achieve a higher degree of robustness to the background changes. The experimental evaluations on a modified MNIST dataset reveals not only that the results are improved but also a higher degree of robustness to the background perturbation using additive noise is obtained.

Iterative feedforward and feedback processes are recognized to play important roles in the information processing of the human brain \cite{gilbert2013top,hupe1998cortical}.
To this end, the Selective Tuning model \cite{tsotsos2011computational,tsotsos1995SelTun,tsotsos2008different} defines multiple computational stages in artificial dynamical networks such as the preliminary stage of visual task priming, the early stage of bottom-up neuronal encoding, the selective stage of top-down attention, and finally the re-interpretation and iterative bottom-up passes. 

\begin{figure*}[]
	\centering
	\includegraphics[width=1.25\columnwidth]{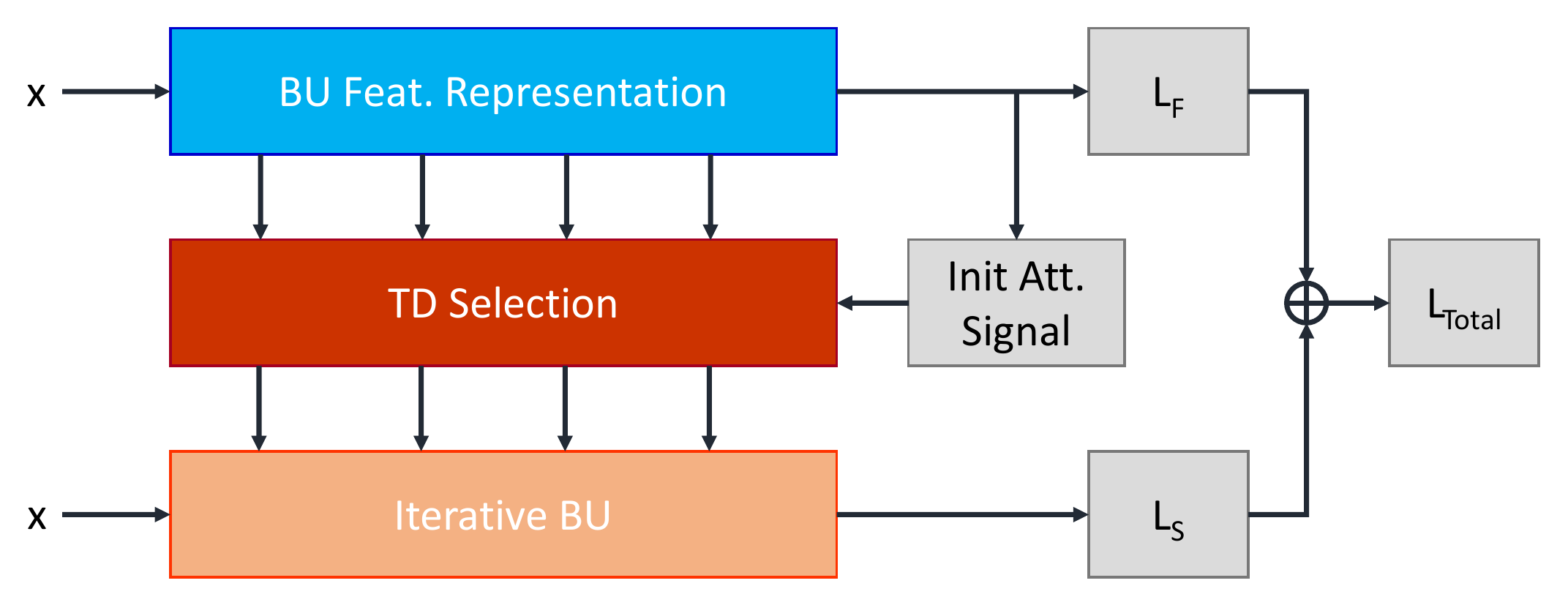}
	\caption{The TD network modulates the BU feature representation in the iterative BU pass. The total loss is defined as the weighted sum of the loss of the first and second BU passes.\label{fig:sft-overall-arch}}
\end{figure*}

Despite the success of Feedforward neural networks in various visual tasks and domains such as object classification \cite{krizhevsky2012imagenet,simonyan2013deep,szegedy2014going,he2016deep,huang2017densely}, object detection \cite{ren2015faster,liu2016ssd,redmon2016yolo9000,he2017mask}, semantic segmentation \cite{long2015fully,chen2016deeplab,islam2017gated,chen2018deeplab}, they still currently suffer from different vulnerabilities
such as visual confusion \cite{biparva2017stnet,rosenfeld2018elephant,dodge2017can}, and adversarial attacks \cite{nguyen2015deep,akhtar2018threat,kurakin2016adversarial} due to the unconstrained and data-driven nature
of the training method in such networks. Semantic objects of unlabeled categories
are confusingly mixed up with the representation of labeled categories. \cite{biparva2017stnet} demonstrates the cases in which the top-down localization leads to the selection of unlabeled object categories with high co-occurrence to labeled categories. Similar types of visual confusion is reported for object detection in \cite{rosenfeld2018elephant}. Human and machine robustness against input distortions is also studied \cite{dodge2017can}. It is revealed that even in the gist representation provided by the feedforward feature encoding, humans are still competent to deal with input noise distortions while neural networks fall behind.
The neural networks, as highly parametric learning machines, are strongly prone to overfitting to the data distribution of the benchmark datasets and consequently achieve low generalization to unseen and distorted data samples.
Addition of extra regularization terms to appropriate objective functions \cite{varga2017gradient} and sparsification of gradients \cite{sun2017meprop,alistarh2018convergence,akhtar2018threat} are two approaches to improve robustness against visual vulnerabilities and generalization performance.

We suggest that implicit concentration of the learning method potential on target
objects can help to reduce the contextual interference in neural networks. Since the spatial extent of objects is gradually lost within the visual hierarchy in neural
networks (the Blurring problem defined in \cite{tsotsos2011computational}), a TD selection mechanism is essential to constraint the
focus of the learning method on relevant spatial regions and feature channels. We hypothesize that training a neural network with iterative BU passes driven from
TD attentive mechanisms will achieve a more robust representation
and improve the localization and categorization prediction metrics. 

STNet \cite{biparva2017stnet} introduces a unified framework with
BU and TD passes. The framework has shown competitive results for tasks such as
object localization.
Building on top of this two-pass framework, we propose a novel iterative framework that benefits from selection patterns generated in the TD pass for the modulation of the feature extraction layers in the iterative BU pass. We show that using a novel multi-loss objective function, the network learns to concentrate the focus of attention on the relevant aspects for feature representation.
This helps the network to escape unreliable local minima in which the localization accuracy is low and the context has been utilized wrongly for label prediction. 
We demonstrate a notion of overfitting when a network is trained to predict category labels while unable to localize objects accurately using the learned representation.
The proposed augmented loss function, derived from the iterative framework, has an implicit regularization impact on the entire learning algorithm.
The experimental evaluation reveals that not only the localization but also classification accuracy rates are improved. The ablation studies demonstrate that the proposed model achieves a higher degree of robustness to the contextual perturbation and hence verifies the attentive capability to focus on relevant encoding aspects.

In Sec. \ref{sec:mod}, the proposed multi-pass and multi-loss processing framework is introduced. Next, The experimental setup using a modified version of MNIST dataset is provided in Sec \ref{sec:exp}. Additionally, the role of the TD gating modulation in the iterative feedforward pass is investigated using several ablation studies.

\begin{figure*}[]
	\centering
	\includegraphics[width=2.0\columnwidth]{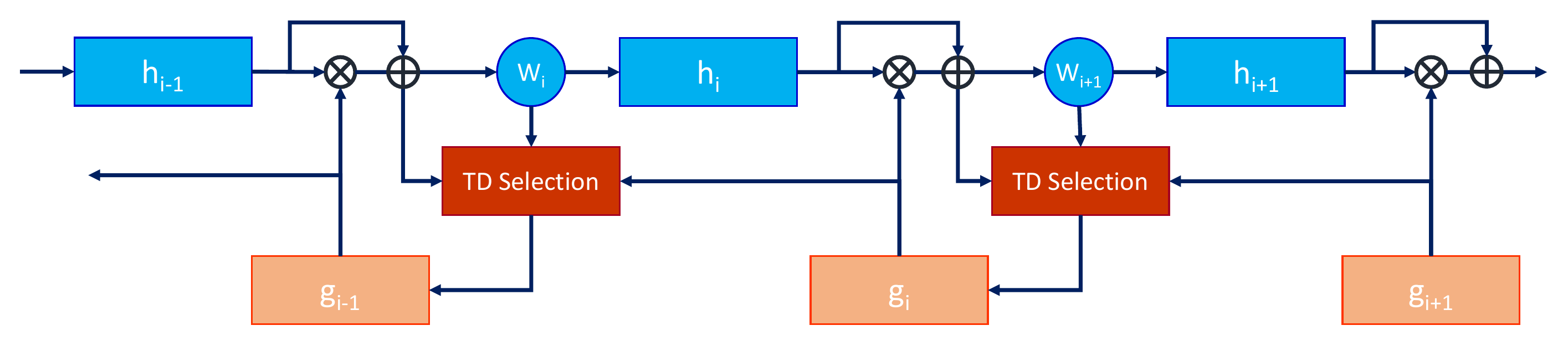}
	\caption{The gating activities at each layer modulate the hidden activities in the iterative BU pass.\label{fig:sft-modular}}
\end{figure*}

%% file: 03-model.tex

We define a neural network framework that consists of the Bottom-Up (BU)
feature representation and the Top-Down (TD) modulatory selection. The BU pass forms the hierarchical feature representation while the TD pass plays the role of a selection mechanism which is capable of gating the subsequent iterative BU pass. In the following, we demonstrate the formulation of each pass and describe the flow of information through the processing pipeline of the multi-pass framework.

\subsection{Initial Feedforward Pass}
The BU network is a regular multi-layer feature extraction model. Having defined a training set $D=\{(x_{i},y_{i})\}_{i=1}^{N}$ of $N$ number
of input image $x\in\mathbb{R}^{H\times W}$ and ground truth category
labels $y\in\{0,\dots,K-1\}$ for $K$ categories, a mini-batch of
training samples are fed into the BU network for category label prediction:

\begin{equation}
s=f(x;W),
\end{equation}
where $x$ is the set of input images, $W$ is the set of BU network
parameters, and $s$ is the output confidence scores of all classes.
After multiple-layers of parametric feature transformation $f$, the
confidence score $s$ is returned to a softmax probability distribution
$p=softmax(s)$ for multinomial category label predictions. $f=\{f_{i}\}_{i=1}^{L}$
is a multi-layer neural network with $L$ layers. It contains the
set of feature transformation functions $f_{i}$ such that $h_{i}=f_{i}(h_{i-1};w_{i})$.
The hidden activities of the previous layer $h_{i-1}$ is the input
and $h_{i}$ is the output of the layer. It is worth mentioning that
$h_{0}=x$ and $h_{L}=s$.

\subsection{Top-Down Selection Pass}
STNet \cite{biparva2017stnet} is a TD processing approach based on the Selective Tuning (ST) computational model of
visual attention \cite{tsotsos1995SelTun,tsotsos2011computational}. We choose to extend STNet as the basis for the TD selection processing that complements the BU processing in a typical convolutional neural network. In both BU and TD passes, the information is traversed according to the flow direction of the pass in a layer-wise manner.

The flow direction is different in each pass, and the previous
and next layers are accordingly defined. In the TD pass, for a layer
$L_{l}$, the previous layer is the top layer $L_{l+1}$ and the next
layer is the bottom layer $L_{l-1}$ within the hierarchy while the
reverse is true for the BU pass. The BU pass benefits from learnable
connection weight parameters for feature transformation such as convolutional
kernels while the TD pass has no such type of weight parameters. Instead,
there are adaptive thresholding rules that specify the selection process
properties. The rules are determined as the result of competitions between
input values at each TD layer. 

Following STNet, the TD pass starts from a top initialization signal and ends at the bottom of the visual hierarchy.
It contains a selection mechanism at every layer consisting of 3 stages
of computation: 1) noise interference reduction, 2) grouping and selection
3) normalization and propagation.

Similar to a convolutional layer, the computation in a TD selection layer is localized over a retrieved receptive field containing the element-wise multiplication of the input hidden activities and the kernel parameters. Given this set of activities to the TD layer, the first stage of computation is proposed to reduce noise interference by pruning redundant activities. The pruning is based on an adaptive thresholding mechanism. The goal of the thresholding mechanism is find the most important subset of activities that participate in the information propagation in the BU pass according to the kernel parameters.
The second stage is defined to impose the connectivity constraint inherently necessary in any reliable visual representation. It groups the activities in the subset of activities returned by the first stage according to the connected-component algorithm, and then selects the group that has the highest value of a combination of the size and total activity strength. The activities in this group participate in the TD propagation of attention signals to the lower layer.
The final stage is defined to normalize the activities of the selected group such that they sum to one, and then propagate the activity of the top gating unit proportional to the normalized activities of the selected group to the gating units of the next layer.
We define the TD network 
\begin{equation}
g=u(d,H,W),
\end{equation}
where $u=\{u_{i}\}_{i=1}^{L}$ is a set of selection layers, $d\in\mathbb{R}^{K}$
is the initialization signal, and $H=\{h_{i}\}_{i=1}^{L}$ is the
set of the BU hidden activities. $d=\delta_{iy}$ is defined using Kronecker
delta. It is a non-zero vector with all elements zero except the one
at the ground truth label $y$. Particularly, at layer $l$, the selection
layer $g_{l-1}=u(g_{l},h_{l-1},w_{l})$ gets the gating activities
$g_{l}$, the hidden activities at the previous layer $h_{l-1}$,
and the kernel filter parameters $w_{l}$. It outputs the gating activities
$g_{l-1}$ at the end of the selection stages.

We try to shift the visual representation of the BU network to concentrate
on the feature channels and spatial regions of the target object in
the foreground rather than the context in the background. Using the
TD pass initialized from the ground truth category labels, the gating
activities at each layer are selective for the subset of features
that are significantly important for the category label predictions.
During the selective fine-tuning phase, the network learns to focus
on the network parameters that are gated by the TD pass.

\begin{table}
	\centering
	\resizebox{0.90\columnwidth}{!}{
		
		\begin{tabular}{l|c|c}
			\hline 
			Model & classification & localization\tabularnewline
			\hline 
			LeNet-5-reference & 94.0\% & 96.4\%\tabularnewline
			LeNet-5-sft & \textbf{97.5\%} & \textbf{99.1\%}\tabularnewline
			\hline 
			AlexNet-reference & 97.1\% & 98.2\%\tabularnewline
			AlexNet-sft & \textbf{99.3\%} & \textbf{99.8\%}\tabularnewline
			\hline 
		\end{tabular}
	}
	\caption{The classification and localization rates of the selective fine-tuned
		network on the WMNIST dataset.\label{tab:sft-metrics}}
\end{table}

\subsection{Iterative Feedforward Pass}

Having defined the feedforward BU and the selective TD passes, we define
the iterative BU pass using the gating activities computed in the
TD pass. For a mini-batch of samples, the BU pass is first activated,
the hidden activities are computed, and the output label prediction
is returned. Next, the initialization signal is set using the ground
truth label, and then the TD pass is triggered to begin. The gating
activities are computed layer by layer until the TD pass stops at the
input layer. Then, we define the iterative BU pass consisting of $L$ layers similar to the initial feedforward pass such that at the layer $i$, the gated hidden activities $t_{i}$ are 

\begin{equation}
	t_{i}=\alpha*\tilde{h}_{i}\odot\tilde{g}_{i}+\beta*\tilde{h}_{i},
\end{equation}
where $a\odot b$ is the Hadamard product of $a$ with $b$, $\tilde{h}_{i}$
is the input hidden activities, and $\tilde{g}_{i}=n(g_{i})$ is the
normalized gating activities using the function $n$ such that $\tilde{g}$
has a minimum and maximum activities of zero and one respectively.
$\alpha$ and $\beta$ are the multiplicative factors to control the
numeric level of the hidden and gating activities respectively. They are set
to one unless otherwise mentioned. Having $t_{i}$ computed, the output
hidden activities at layer $i+1$ is computed 

\begin{equation}
	\tilde{h}_{i+1}=f(t_{i};w_{i+1}).
\end{equation}

Using the confidence score output $\tilde{s}=f(x;W)$, the multinomial probability
prediction of the iterative pass is $\tilde{p}=softmax(\tilde{s})$. 
We propose an attention-augmented loss function with two terms $\mathcal{L}_{F}$ and $\mathcal{L}_{S}$:

\begin{equation}
\label{eq:ch7:total_loss}
	\mathcal{L}_{T}(p,\tilde{p},y)=\frac{1}{N}\sum_{i}\mathcal{L}_{F}(p_{i},y_{i})+\alpha\frac{1}{N}\sum_{i}\mathcal{L}_{S}(\tilde{p_{i}},y_{i}),
\end{equation}
where $p$ and $\tilde{p}$ are the class probabilities using the
first and iterative BU passes respectively, and $y$ is the ground
truth class label. $\alpha$ is the factor that defines the emphasis
on either term. It is set to one unless otherwise stated. 
$\mathcal{L}_{F}$ and $\mathcal{L}_{S}$ are the cross-entropy loss functions for the true target labels $y_{i}$ and the probability predictions $p_{i}$ and $\tilde{p_{i}}$ of the first and iterative feedforward passes respectively. 
The cross entropy loss function $\hat{\mathcal{L}}$ is defined as:

\begin{equation}
	\hat{\mathcal{L}}( p_{i},y_{i} ) = - \sum^N_{i=1} \sum^K_{k=1} 1 \{ y_{(i)} = k \} \log p(y_{(i)} =  k \mid x_{(i)} ; W),
\end{equation}
where the indicator function $1\{a=b\}$ is one if $a=b$ and zero otherwise. $p(y_{(i)} =  k \mid x_{(i)} ; W) = p^k_i$ is the softmax prediction probability of class $k$ given the input sample $x_i$ and the network parameter $W$.
The first term in the definition of $\mathcal{L}_{T}$ maintains the representational fidelity to the pre-trained BU network while the second term enforces the concentration of the learning algorithm on the TD attention traces. This encourages the network to learn to separate the representations of the background context from the foreground target objects. 
This hypothesis is examined in experimental evaluation and the observations supporting the role of attention to untangle the representation are demonstrated in Sec. \ref{sec:exp}.

\begin{figure}[t]
	\includegraphics[width=0.25\columnwidth]{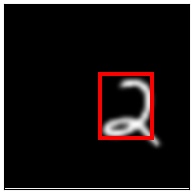}\includegraphics[width=0.25\columnwidth]{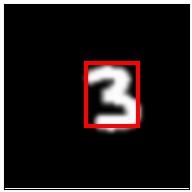}\includegraphics[width=0.25\columnwidth]{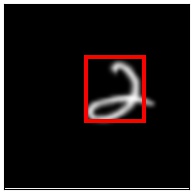}\includegraphics[width=0.25\columnwidth]{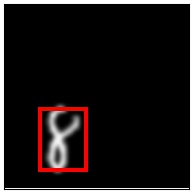}
	
	\includegraphics[width=0.25\columnwidth]{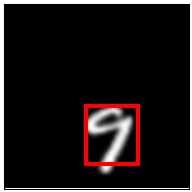}\includegraphics[width=0.25\columnwidth]{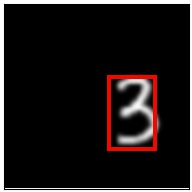}\includegraphics[width=0.25\columnwidth]{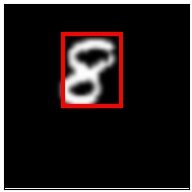}\includegraphics[width=0.25\columnwidth]{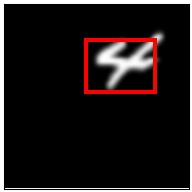}
	
	\includegraphics[width=0.25\columnwidth]{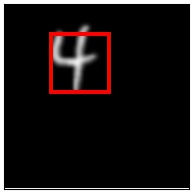}\includegraphics[width=0.25\columnwidth]{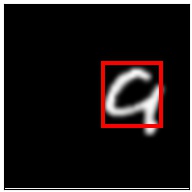}\includegraphics[width=0.25\columnwidth]{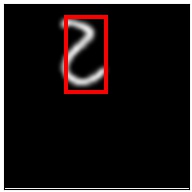}\includegraphics[width=0.25\columnwidth]{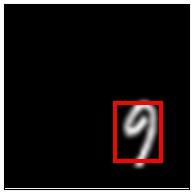}
	
	\caption{Illustration of sample digit images in the WMNIST dataset. The red boxes are the predicted bounding boxes using the LeNet-5 BU pass for feature encoding and the TD selection pass for object localization.\label{fig:Illustration-of-digit}}
\end{figure}

The Stochastic Gradient Descent (SGD) optimization method is used for the training of the neural
network. The error gradients are computed using the loss function
and propagated backward to the input layer. The weight gradients are
accumulated using the computation graphs generated in the first and
iterative BU passes. They each contribute separately to the accumulation
of gradients to update weight parameters at each SGD updating iteration.
Importantly, the error gradients through the iterative BU pass are
back-propagated according to the gating patterns that impacted the feedforward
information flow in the iterative BU pass. This gating mechanism helps the
optimization algorithm focus on the spatial regions and feature channels
that most contributed to the prediction of the input samples at the
first pass. 
The gradient signals are masked at each layer according to the selection patterns formed by the gating activities.
Over various updating iterations, the network learns the representation using which a higher degree of robustness to contextual perturbation is obtained.

\begin{figure}[t]
	\centering
	\includegraphics[width=0.75\columnwidth]{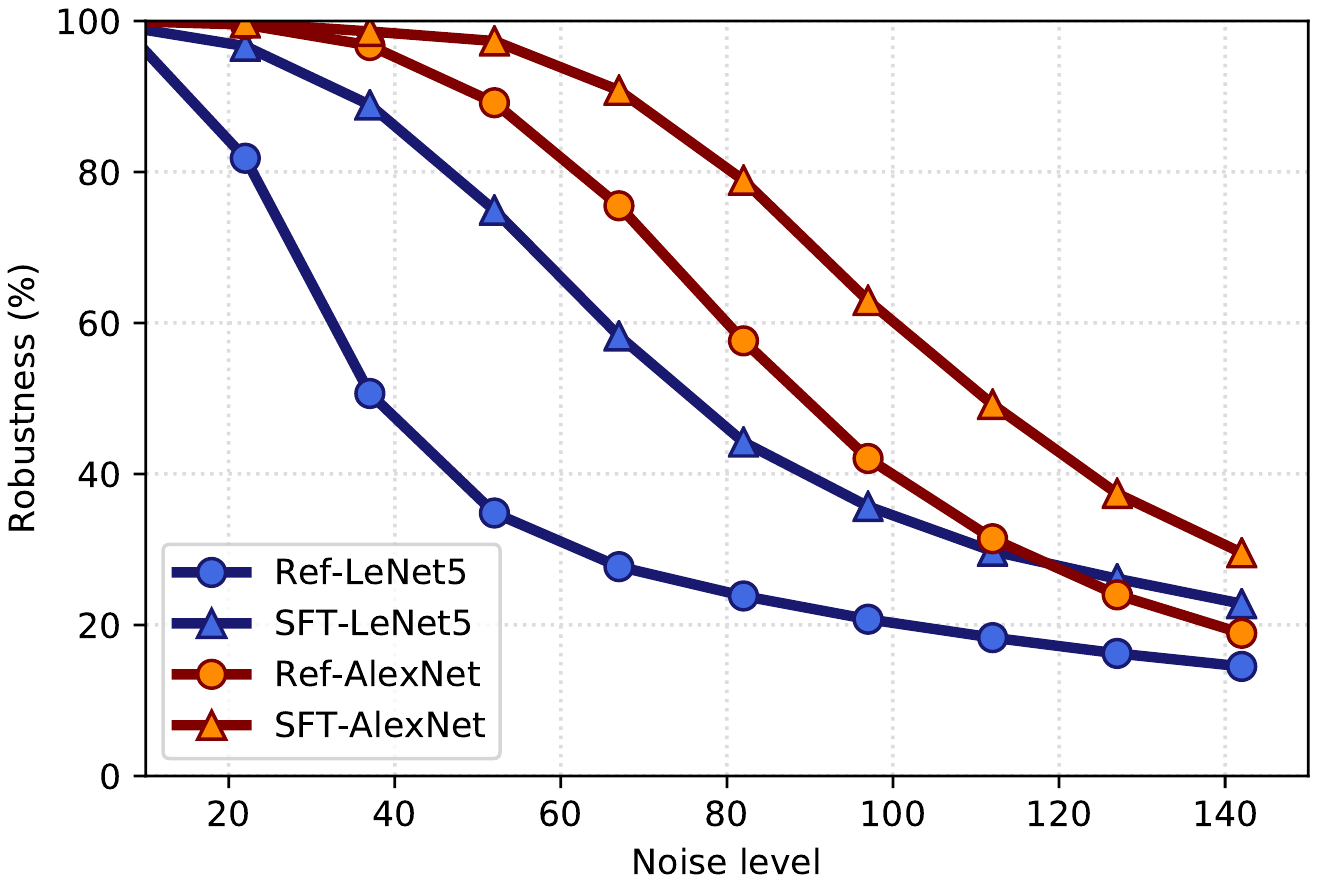}
	
	\caption{The effect of the additive noise distortion in the background on the classification accuracy rate. Ref and SFT refer to the reference and selectively fine-tuned models respectively. The vertical axis represents the robustness of the fine-tuned network at different noise levels. Robustness is calculated by the ratio of the accuracy rates of the noisy images over the clean images. The horizontal axis indicates the maximum amount of pixel intensity the uniform distribution may add to the background pixels. \label{fig:cls}}
\end{figure}

Figure \ref{fig:sft-overall-arch} depicts the flow of the information from the BU feature representation into the TD selective attention block. Once the TD pass ends at the end of the visual hierarchy, the iterative BU pass is started given the same mini-batch of input data. The iterative feedforward pass has modulatory units that change the information flow according to the gating activity responses. The iterative pass, therefore, forms a visual representation with an emphasis on the attended regions and feature channels. The confidence score outputs of the two feedforward passes define the $\mathcal{L}_{F}$ and $\mathcal{L}_{S}$ loss terms that are combined in Eq. \ref{eq:ch7:total_loss} to define the total loss function $\mathcal{L}_{T}$. Once the loss value is computed, the computation graph is used in the SGD optimization algorithm to calculate the parameter gradients of the entire network. The SGD optimization algorithm aims to minimize  $\mathcal{L}_{T}$ in the fine-tuning phase. This basically means that the negative log-likelihood functions derived from the confidence score outputs at the end of the two feedforward passes needs to be reduced. This further implies that the learned representation needs to maintain the class probability prediction capability at a high level of accuracy in the two feedforward passes. Not only does the first feedforward pass is important similar to a regular fine-tuning approach, but also the emphasis to the important aspects of the learned representation is increased by the attentive TD gating mechanisms in the iterative feed forward pass.
Fig. \ref{fig:sft-modular} provides in detail the information flow in the BU pass, the TD pass, and the modulatory interaction of the TD pass with the iterative BU pass. At each layer, the gating activities $g_i$ modulates the hidden activities $h_i$ in the second BU pass. The result then is passed to the parametric transformation function.

%% file: 04-experiments.tex
\begin{figure}[t]
	\includegraphics[width=0.165\columnwidth]{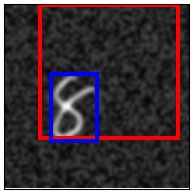}\includegraphics[width=0.165\columnwidth]{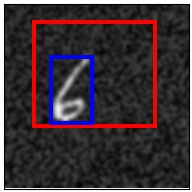}\includegraphics[width=0.165\columnwidth]{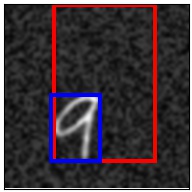}\includegraphics[width=0.165\columnwidth]{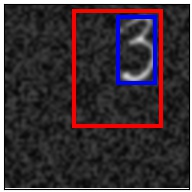}\includegraphics[width=0.165\columnwidth]{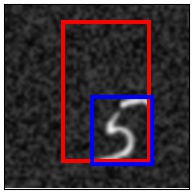}\includegraphics[width=0.165\columnwidth]{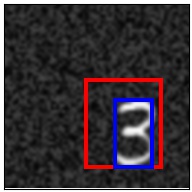}
	
	\includegraphics[width=0.165\columnwidth]{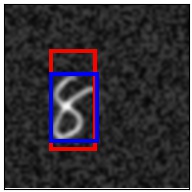}\includegraphics[width=0.165\columnwidth]{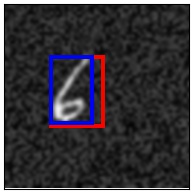}\includegraphics[width=0.165\columnwidth]{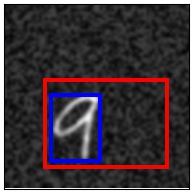}\includegraphics[width=0.165\columnwidth]{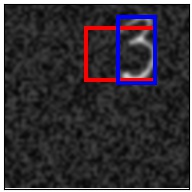}\includegraphics[width=0.165\columnwidth]{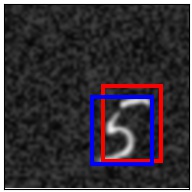}\includegraphics[width=0.165\columnwidth]{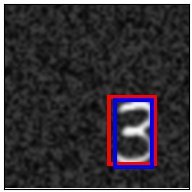}
	
	\caption{Demonstration of the effect of the additive uniform noise in the background and the comparison of the localization performance of the the LeNet-5 reference model (top) with the selective fine-tuned model (bottom). 		The ground truth and predicted boxes are depicted with the blue and red boxes respectively. The additive noise is taken from a uniform distribution with a lower and upper bounds of 0 and 100 respectively.\label{fig:noise}}
\end{figure}

We evaluate the proposed selective fine-tuning of neural networks
on a modified MNIST dataset called Wide-MNITS (WMNIST). MNIST is a
handwritten digit classification dataset. The gray-scale image samples
in the dataset contain handwritten digits of category zero to nine.
We pre-train the BU network on WMNIST for 15 epochs before the evaluation
of the proposed method. Once, the BU network is selectively fine-tuned
for a number of epochs, we measure the robustness of the final
network to the background noise perturbation. The experimental results
reveal that the attention-augmented loss function improves the accuracy
rate while obtain stronger robustness to noise perturbation.

\subsection{Implementation Details}

We define two choices of convolutional neural network architecture for the BU network:
LeNet-5 \cite{Lecun1998} or AlexNet \cite{krizhevsky2012imagenet}.
The TD network is defined by extending the implementation of STNet\footnote{\href{https://github.com/mbiparva/stnet-object-localization}{https://github.com/mbiparva/stnet-object-localization}} \cite{biparva2017stnet} for object localization to consider the new requirements of the iterative BU pass. We define the BU and TD framework in PyTorch deep learning framework \footnote{\href{https://pytorch.org/}{https://pytorch.org/}}
\cite{paszke2017automatic}. The dynamic graph engine in Pytorch allows
the active gating of the hidden activities in the iterative pass to be systematically implemented. The SGD optimization
method uses the learning rate $10^{-3}$, momentum $0.9$, weight
decay $0.0005$, and mini-batch size $64$ unless otherwise mentioned for the pre-training and fine-tuning phases. 
Having the pre-trained BU network loaded, using the selective fine-tuning
method, we update the network parameters for 15 epochs and then report
the accuracy metric in Table \ref{tab:sft-metrics}.

\begin{figure}
	\centering
	\includegraphics[width=0.75\columnwidth]{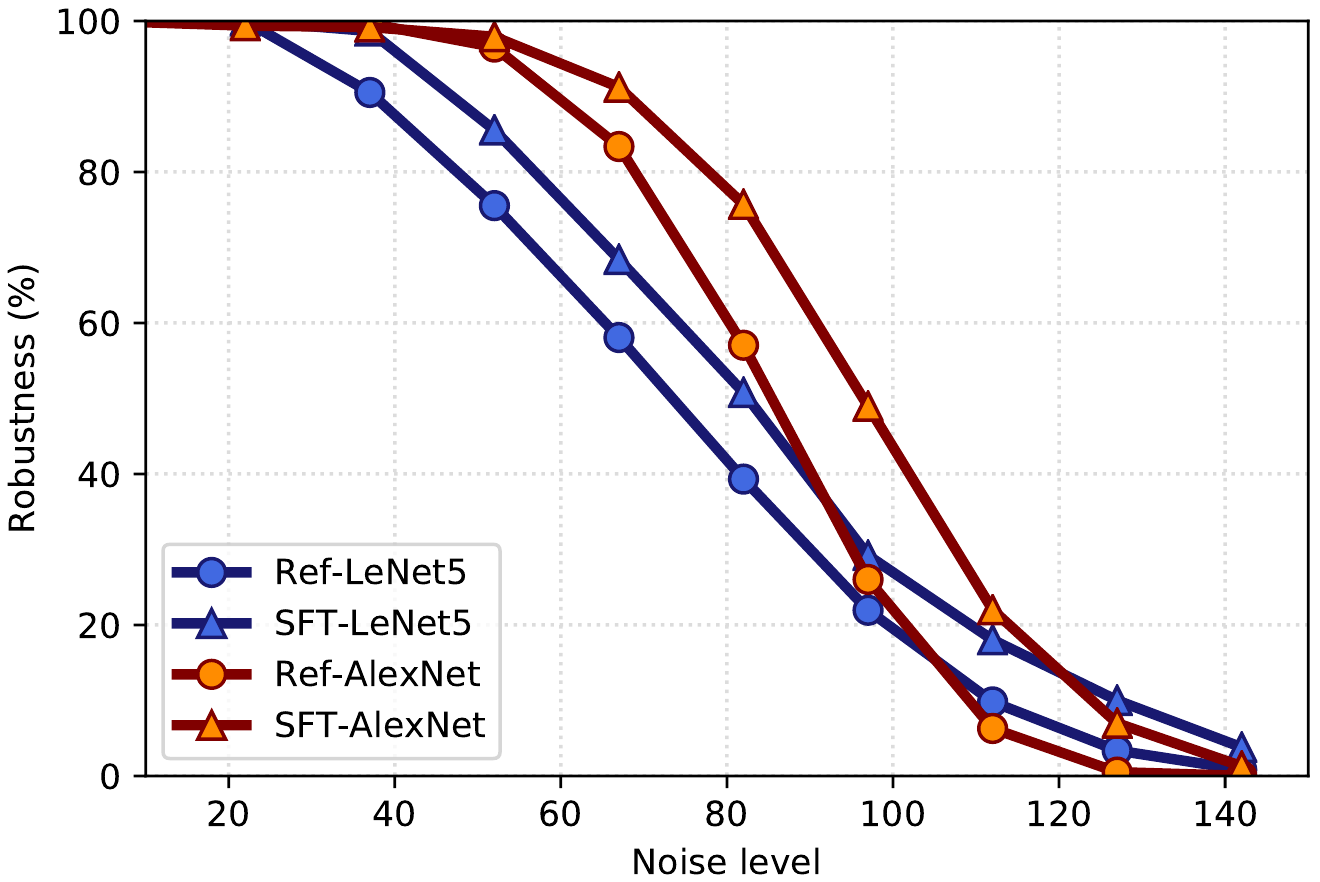}
	
	\caption{The effect of the additive noise in the background on the localization
		accuracy rate. Ref and SFT refer to the reference and selective fine-tuned
		models respectively. The horizontal axis indicates the maximum amount
		of pixel intensity the uniform distribution may add to the background
		pixels. Robustness is calculated by the ratio of the accuracy rates of the noisy images over the clean images.\label{fig:loc}}
\end{figure}

\subsection{Wide MNIST Dataset}

The experimental evaluation is designed to examine the role of the background
context for the category label prediction of the foreground target
object. The role of the background representation is explicitly highlighted
by considering a relatively large context in the input data distribution.

\textbf{Dataset and Evaluation:} MNIST dataset contains $28\times28$
gray-scale digit images. We increase the size of images
by expanding the background context such that images have the size
$64\times64$. We additionally randomize the location of digits in
images. In addition to the ground truth labels, while expanding image
samples based on the aforementioned protocol, we also extract the
tightest bounding box around the digit shape. We use both types of
ground truth to measure the performance of the proposed method using
the 0-1 classification and the IoU (0.5) localization accuracy rates.

\textbf{Quantitative Results:} The evaluation result for the LeNet-5
and AlexNet using the classification and localization metrics are
reported in Table \ref{tab:sft-metrics}. The selectively fine-tuned
neural networks report improved performance results. The results underline the role of the TD selective pass on network parameter optimization using the attention-augmented loss function. Not only the localization but also the classification results are improved once the network is fine-tuned using the proposed approach. Fig. \ref{fig:Illustration-of-digit} illustrates sample
images with the predicted bounding boxes as the means of object localization using the LeNet-5 network architecture.
The bounding boxes are predicted using the localization approach presented in STNet \cite{biparva2017stnet}. Since the gating activities at the input
layer are used for box predictions and the input images are gray-scale,
we only need to find a tight enclosing box around all of the non-zero
gating units. We do not use any pruning strategy to remove units with small gating values.

\begin{figure}
	\centering
	\subfloat[Grating]{
		\includegraphics[width=0.20\columnwidth]{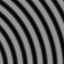}\hspace{0.01\columnwidth}
		\includegraphics[width=0.20\columnwidth]{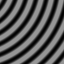}\hspace{0.01\columnwidth}
		\includegraphics[width=0.20\columnwidth]{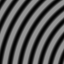}\hspace{0.01\columnwidth}
		\includegraphics[width=0.20\columnwidth]{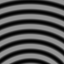}							
	}\\
	\subfloat[MoG]{
		\includegraphics[width=0.20\columnwidth]{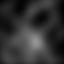}\hspace{0.01\columnwidth}
		\includegraphics[width=0.20\columnwidth]{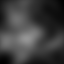}\hspace{0.01\columnwidth}
		\includegraphics[width=0.20\columnwidth]{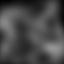}\hspace{0.01\columnwidth}
		\includegraphics[width=0.20\columnwidth]{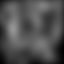}			
	}\\
	\subfloat[Squares]{
		\includegraphics[width=0.20\columnwidth]{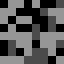}\hspace{0.01\columnwidth}
		\includegraphics[width=0.20\columnwidth]{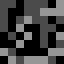}\hspace{0.01\columnwidth}
		\includegraphics[width=0.20\columnwidth]{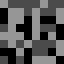}\hspace{0.01\columnwidth}
		\includegraphics[width=0.20\columnwidth]{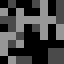}		
	}\\
	\subfloat[RLines]{
		\includegraphics[width=0.20\columnwidth]{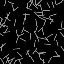}\hspace{0.01\columnwidth}
		\includegraphics[width=0.20\columnwidth]{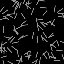}\hspace{0.01\columnwidth}
		\includegraphics[width=0.20\columnwidth]{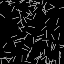}\hspace{0.01\columnwidth}
		\includegraphics[width=0.20\columnwidth]{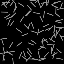}		
		
	}\caption{\label{fig:sft:multi-noise:samples}Random samples generated by the four noise methods: (a) Grating: radial grating with random centers, (b) MoG: Mixture of Gaussians, (c) Squares: squares with random intensity values, and (d) RLines: short lines with random centers and orientation.}
	
\end{figure}

\begin{figure*}
	\centering
	\includegraphics[width=0.10\columnwidth]{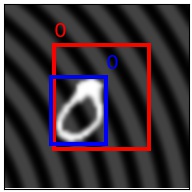}
	\includegraphics[width=0.10\columnwidth]{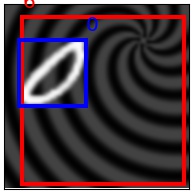}
	\includegraphics[width=0.10\columnwidth]{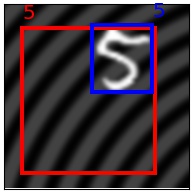}
	\includegraphics[width=0.10\columnwidth]{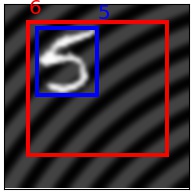}
	\includegraphics[width=0.10\columnwidth]{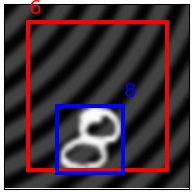}
	\includegraphics[width=0.10\columnwidth]{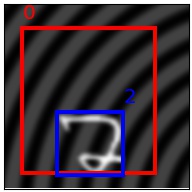}
	\includegraphics[width=0.10\columnwidth]{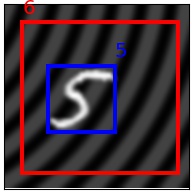}
	\includegraphics[width=0.10\columnwidth]{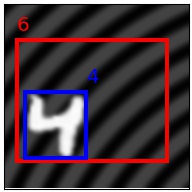}
	\vspace{0.005\columnwidth}
	\includegraphics[width=0.10\columnwidth]{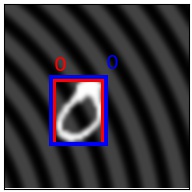}
	\includegraphics[width=0.10\columnwidth]{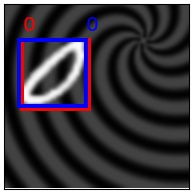}
	\includegraphics[width=0.10\columnwidth]{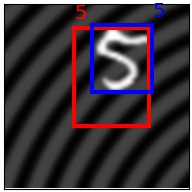}
	\includegraphics[width=0.10\columnwidth]{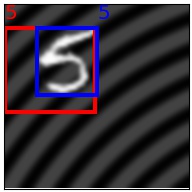}
	\includegraphics[width=0.10\columnwidth]{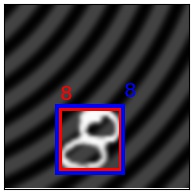}
	\includegraphics[width=0.10\columnwidth]{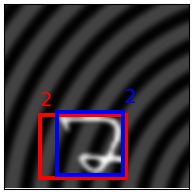}
	\includegraphics[width=0.10\columnwidth]{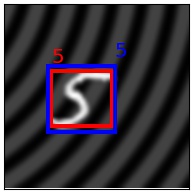}
	\includegraphics[width=0.10\columnwidth]{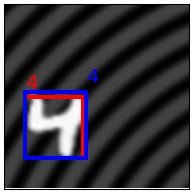}\\		
	{\scriptsize (a) Grating}
	\vspace{0.01\columnwidth}
	
	\includegraphics[width=0.10\columnwidth]{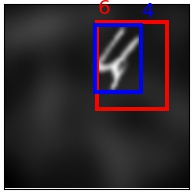}
	\includegraphics[width=0.10\columnwidth]{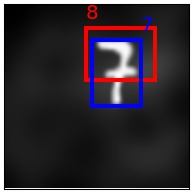}
	\includegraphics[width=0.10\columnwidth]{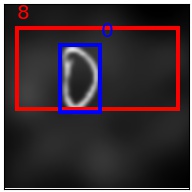}
	\includegraphics[width=0.10\columnwidth]{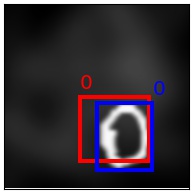}
	\includegraphics[width=0.10\columnwidth]{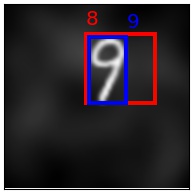}
	\includegraphics[width=0.10\columnwidth]{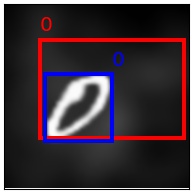}
	\includegraphics[width=0.10\columnwidth]{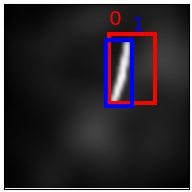}
	\includegraphics[width=0.10\columnwidth]{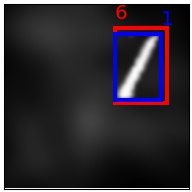}
	\vspace{0.005\columnwidth}
	\includegraphics[width=0.10\columnwidth]{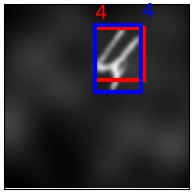}
	\includegraphics[width=0.10\columnwidth]{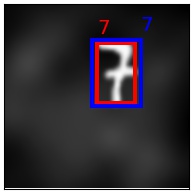}
	\includegraphics[width=0.10\columnwidth]{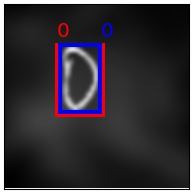}
	\includegraphics[width=0.10\columnwidth]{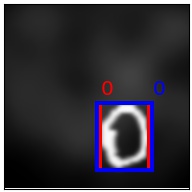}
	\includegraphics[width=0.10\columnwidth]{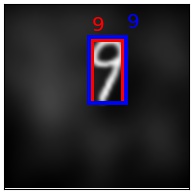}
	\includegraphics[width=0.10\columnwidth]{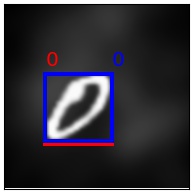}
	\includegraphics[width=0.10\columnwidth]{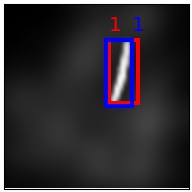}
	\includegraphics[width=0.10\columnwidth]{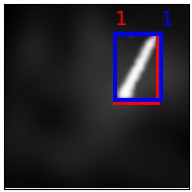}\\		
	{\scriptsize (b) MoG}
	\vspace{0.01\columnwidth}
	
	\includegraphics[width=0.10\columnwidth]{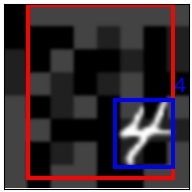}
	\includegraphics[width=0.10\columnwidth]{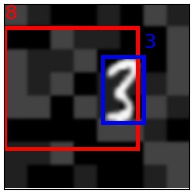}
	\includegraphics[width=0.10\columnwidth]{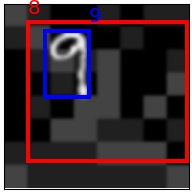}
	\includegraphics[width=0.10\columnwidth]{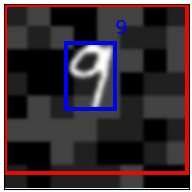}
	\includegraphics[width=0.10\columnwidth]{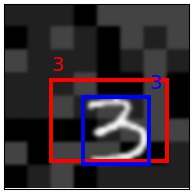}
	\includegraphics[width=0.10\columnwidth]{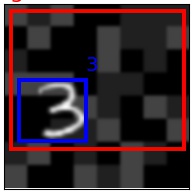}
	\includegraphics[width=0.10\columnwidth]{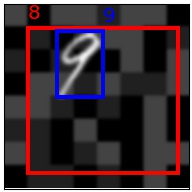}
	\includegraphics[width=0.10\columnwidth]{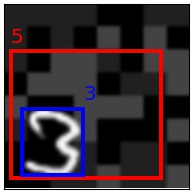}
	\vspace{0.005\columnwidth}
	\includegraphics[width=0.10\columnwidth]{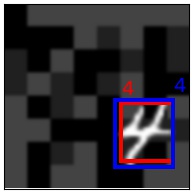}
	\includegraphics[width=0.10\columnwidth]{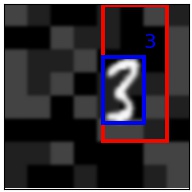}
	\includegraphics[width=0.10\columnwidth]{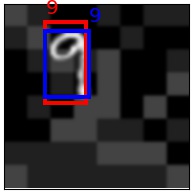}
	\includegraphics[width=0.10\columnwidth]{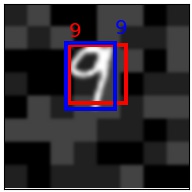}
	\includegraphics[width=0.10\columnwidth]{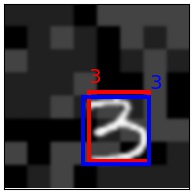}
	\includegraphics[width=0.10\columnwidth]{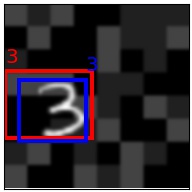}
	\includegraphics[width=0.10\columnwidth]{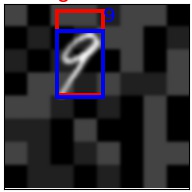}
	\includegraphics[width=0.10\columnwidth]{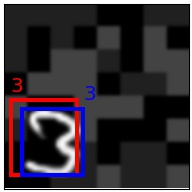}\\		
	{\scriptsize (d) Squares}
	\vspace{0.01\columnwidth}		
	
	\includegraphics[width=0.10\columnwidth]{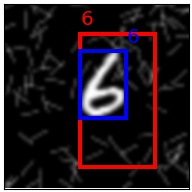}
	\includegraphics[width=0.10\columnwidth]{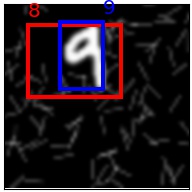}
	\includegraphics[width=0.10\columnwidth]{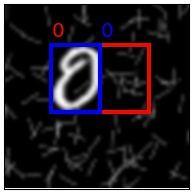}
	\includegraphics[width=0.10\columnwidth]{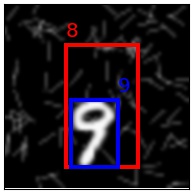}
	\includegraphics[width=0.10\columnwidth]{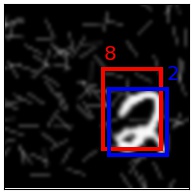}
	\includegraphics[width=0.10\columnwidth]{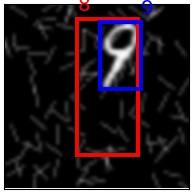}
	\includegraphics[width=0.10\columnwidth]{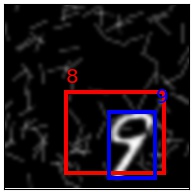}
	\includegraphics[width=0.10\columnwidth]{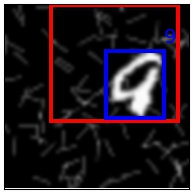}
	\vspace{0.005\columnwidth}
	\includegraphics[width=0.10\columnwidth]{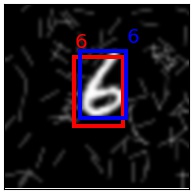}
	\includegraphics[width=0.10\columnwidth]{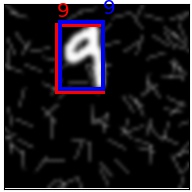}
	\includegraphics[width=0.10\columnwidth]{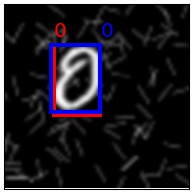}
	\includegraphics[width=0.10\columnwidth]{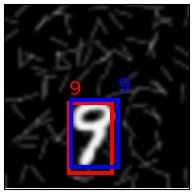}
	\includegraphics[width=0.10\columnwidth]{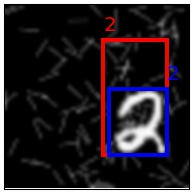}
	\includegraphics[width=0.10\columnwidth]{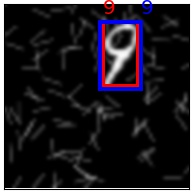}
	\includegraphics[width=0.10\columnwidth]{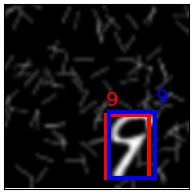}
	\includegraphics[width=0.10\columnwidth]{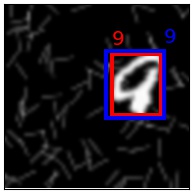}\\		
	{\scriptsize (c) RLines}
	\vspace{0.01\columnwidth}
	
	\caption{\label{fig:sft:multi-noise:lenet5:pred}Comparison of the label and bounding box predictions of the LeNet-5 reference and fine-tuned networks once the background regions is perturbed with four different types of noise methods. In each section, the top and bottom rows represent predictions from the reference and selective fine-tuned networks. The ground truth and predicted bounding boxes are illustrated with blue and red boxes respectively. The ground truth and predicted labels are shown at the top-left and top-right of their corresponding box respectively.}
	
\end{figure*}

\textbf{Ablation Analysis:} We study further the role of the selective
fine-tuning method on the separation of the foreground from the background
representations. We use additive uniform noise in the background to
study the impact of the context interference on the target object
classification and localization predictions. We gradually increase
the upper bound of the uniform noise function to measure the robustness
of the reference and fine-tuned models in sever situations. Fig. \ref{fig:cls}
demonstrates the amount of classification robustness obtained using the selective
models over the reference models for different levels of background additive noise. For both LeNet-5 and AlexNet network architectures, the selective fine-tuning brings a significant level of robustness to the reference models. This result indicates that during selective fine-tuning, the network learns to focus further on the features encoding of the foreground target objects and blocking contextual interference. In addition to the classification task, Fig. \ref{fig:loc} reveals the
localization accuracy is also maintained over different levels of additive noise using the proposed method. Fig. \ref{fig:noise}
qualitatively illustrates the cases the reference model fails to deal
with the background noise. It underlines the fact that in the reference
model the representation of the background context is entangled with the
foreground target object. This explains why a simple form
of contextual perturbation quickly destroys the localization and classification
performance of the reference model. The selective fine-tuning approach,
however, obtains a higher degree of robustness in such sever cases.
Apparently, the network learns through the selectivity of the TD attention to concentrate on the foreground representation.

\begin{figure}
	\centering
	\subfloat[Grating]{\includegraphics[width=0.25\columnwidth]{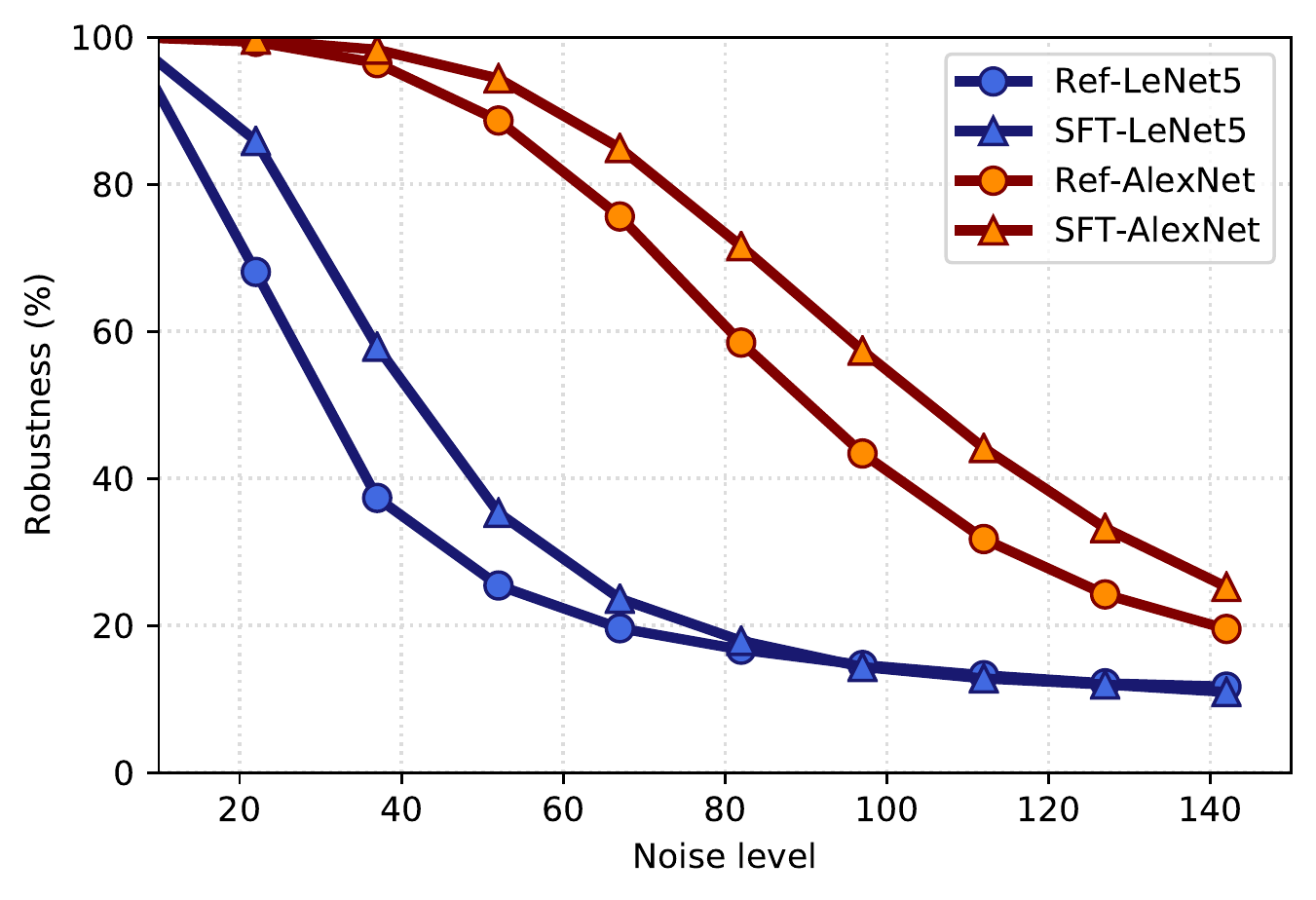}
		
	}\subfloat[MoG]{\includegraphics[width=0.25\columnwidth]{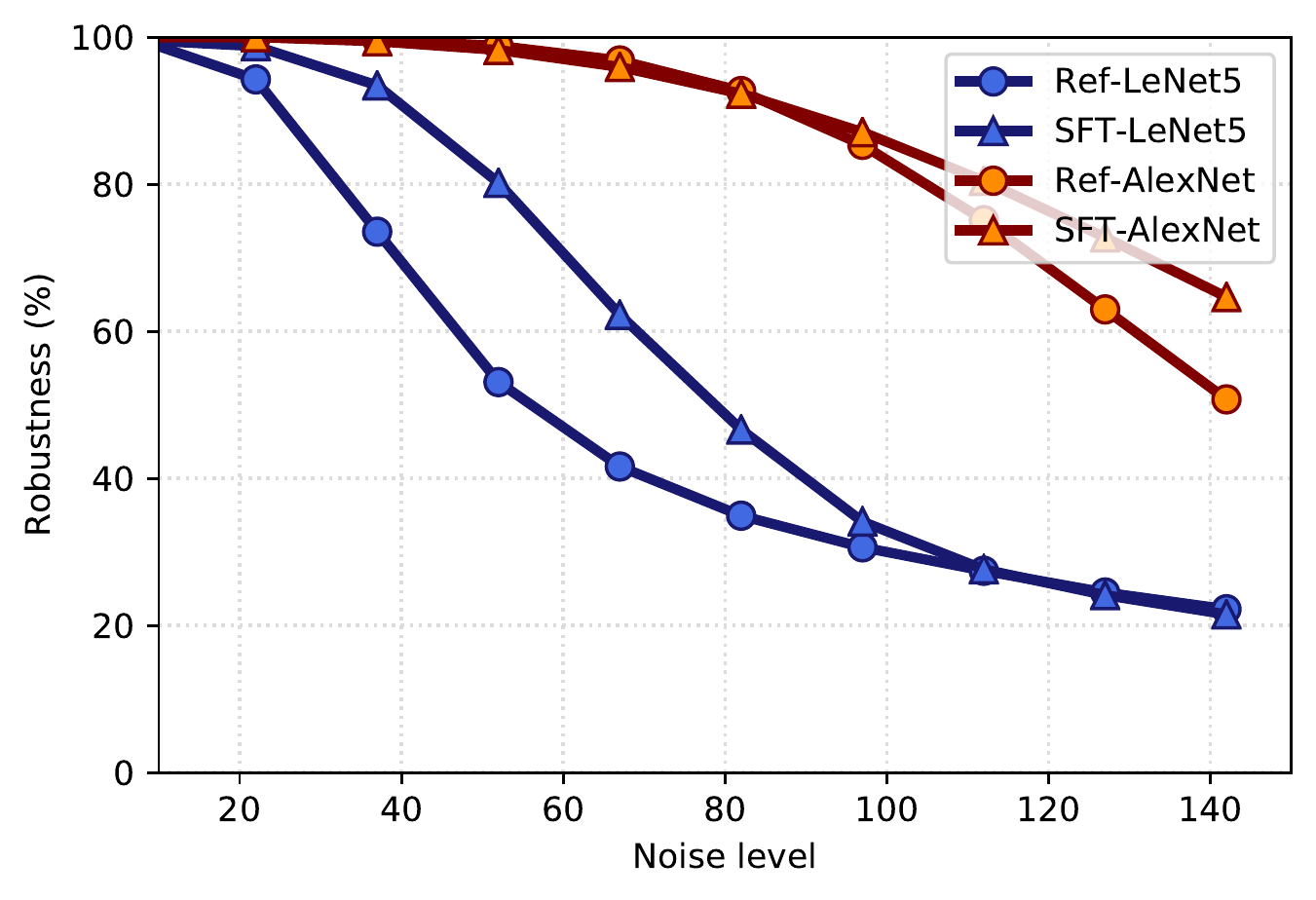}
		
	}\subfloat[Squares]{\includegraphics[width=0.25\columnwidth]{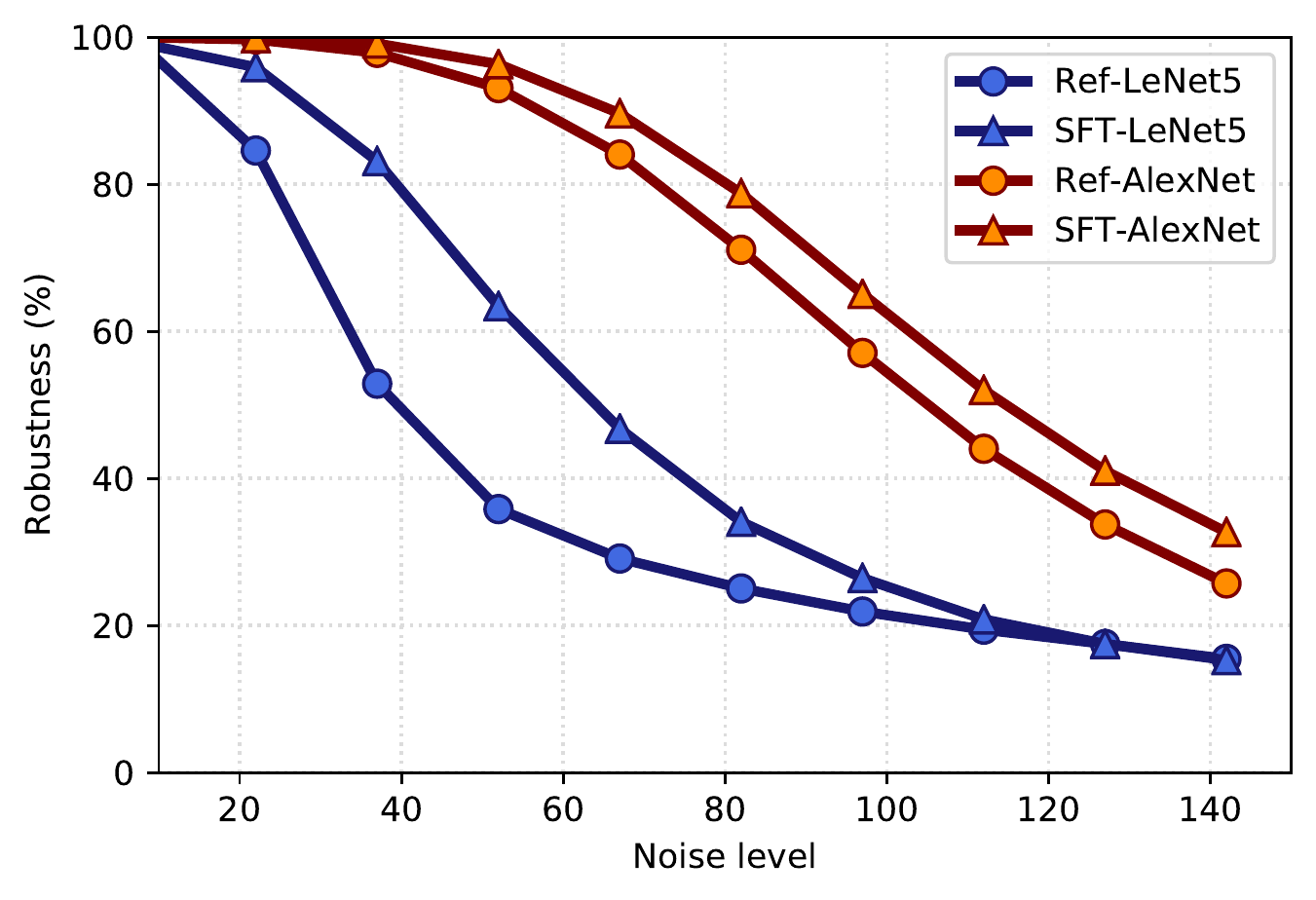}
		
	}\subfloat[RLines]{\includegraphics[width=0.25\columnwidth]{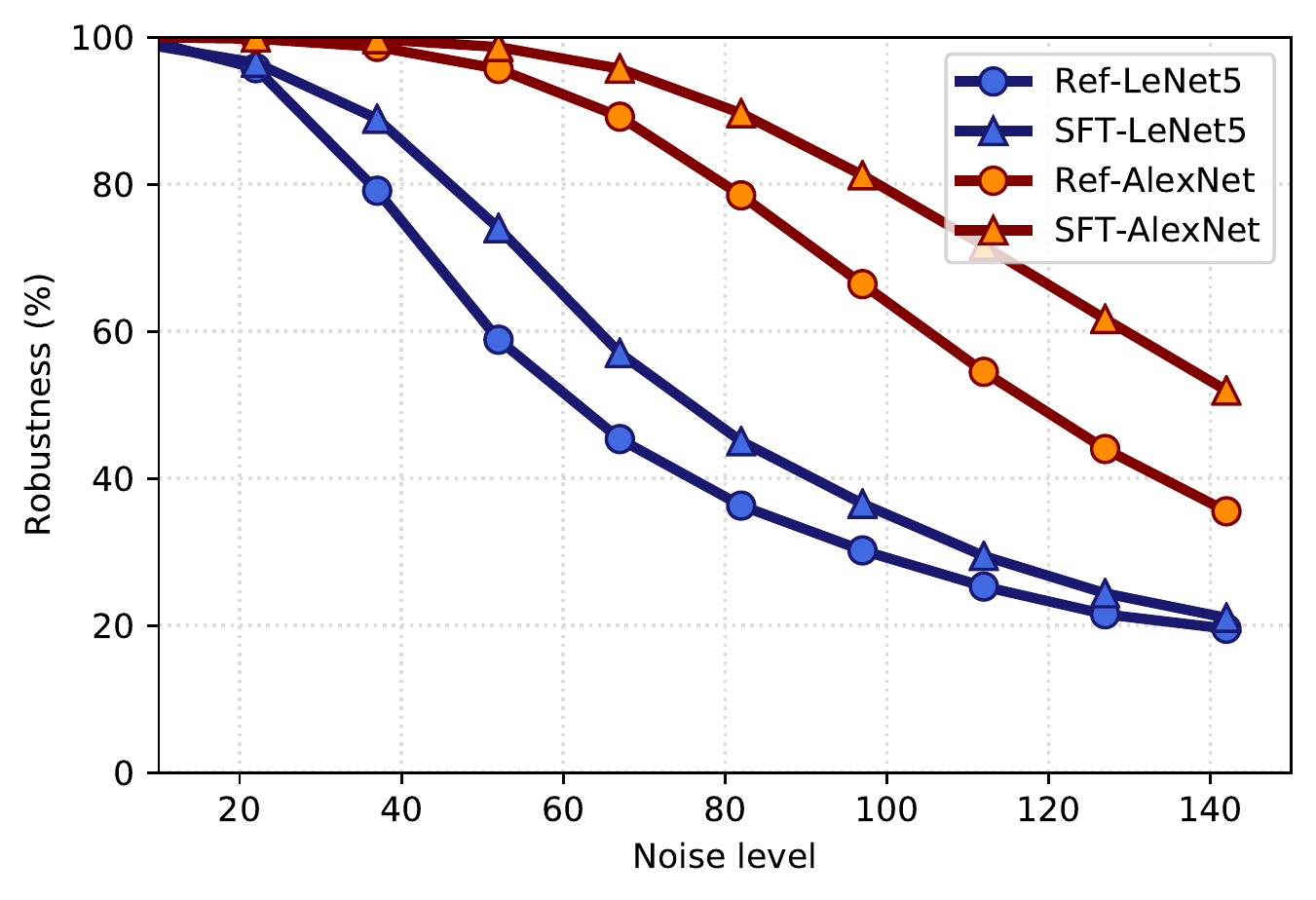}
		
	}\caption{\label{fig:sft:multi-noise:cls}Comparing the effect of different methods of generating contextual noise perturbation on the classification accuracy. From left to right: (a) Grating: radial grating with random centers, (b) MoG: Mixture of Gaussians, (c) Squares: squares with random intensity values, and (d) RLines: short lines with random centers and orientation. The vertical axis represent the classification robustness metric, and the horizontal axis represent the maximum pixel intensity the noise adds to the background.}
	
\end{figure}

We further experiment with different types of noise generation functions to validate the generalization achieved by the selective fine-tuning approach. We choose four different noise sources based on which we choose to perturb the background regions as follows: (1) Grating: this is the radial grating method with a center coordinate randomly chosen for every input image. (2) MoG: this is a mixture of K Gaussian distributions such that each Gaussian has a random center coordinate and orientation. We choose K=50 since it provides smooth and irregular noise patterns. (3) Squares: this generates a K$\times$K grid of squares with random pixel intensity values. We choose K=8 since it generates large enough square blocks that distinguishes them from random uniform noise patterns. (4) RLines: this generates K short line segments with random center coordinates and orientations. We choose K=100 to cover the entire background regions with enough number of noise patterns. Figure \ref{fig:sft:multi-noise:samples} illustrates four random samples generated by these noise generation methods. 

\begin{figure}
	\centering
	\subfloat[Grating]{\includegraphics[width=0.25\columnwidth]{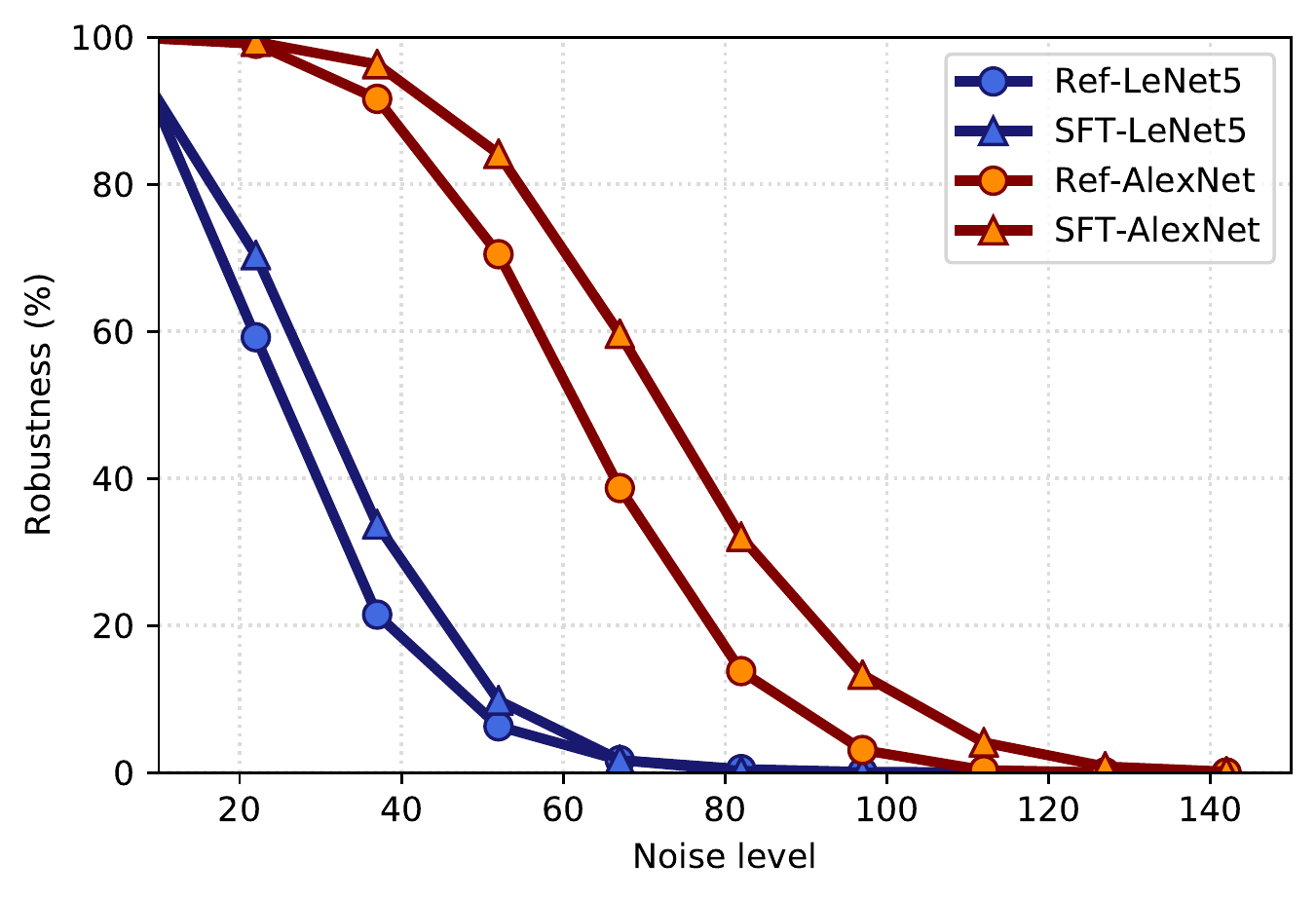}
		
	}\subfloat[MoG]{\includegraphics[width=0.25\columnwidth]{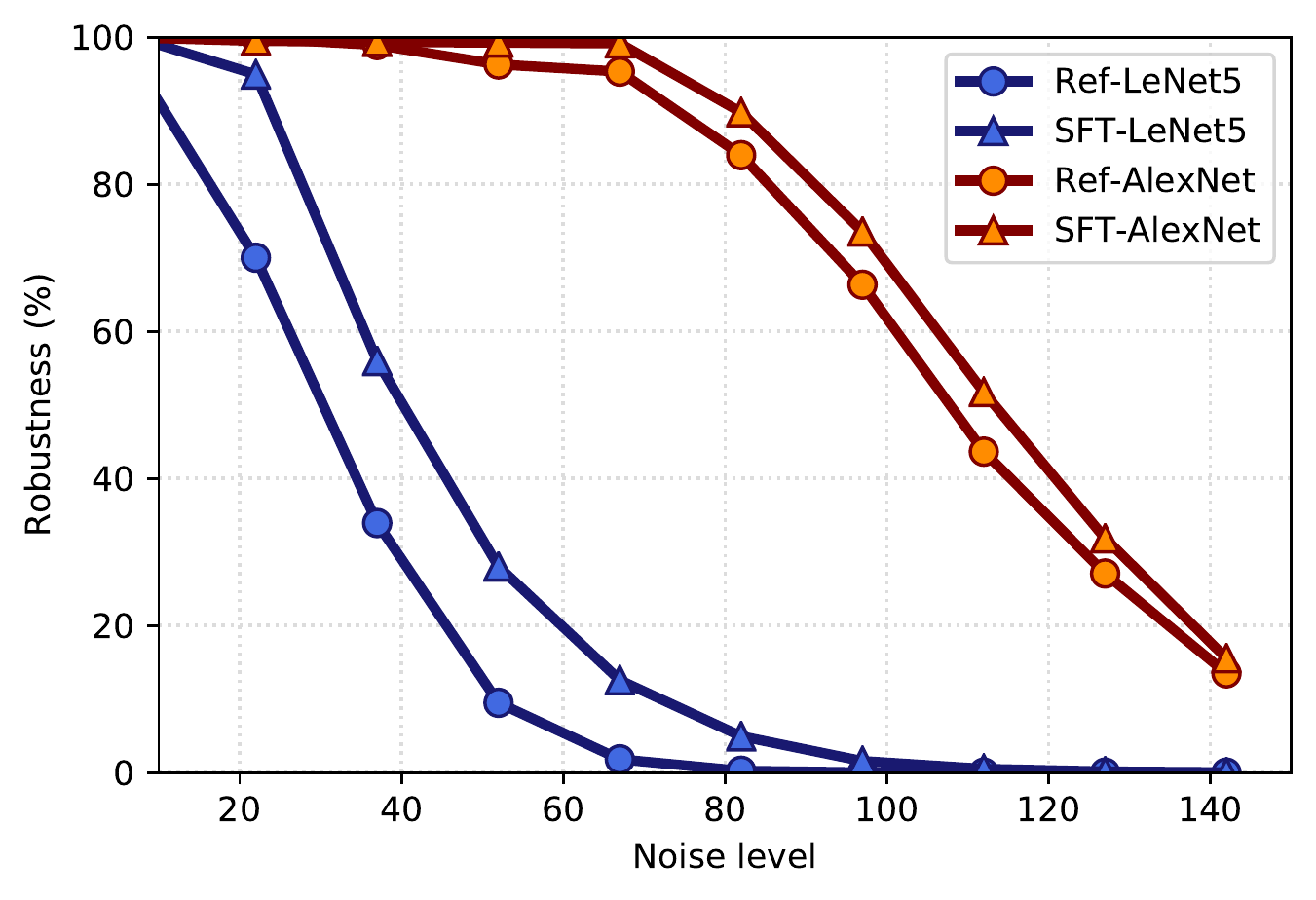}
		
	}\subfloat[Squares]{\includegraphics[width=0.25\columnwidth]{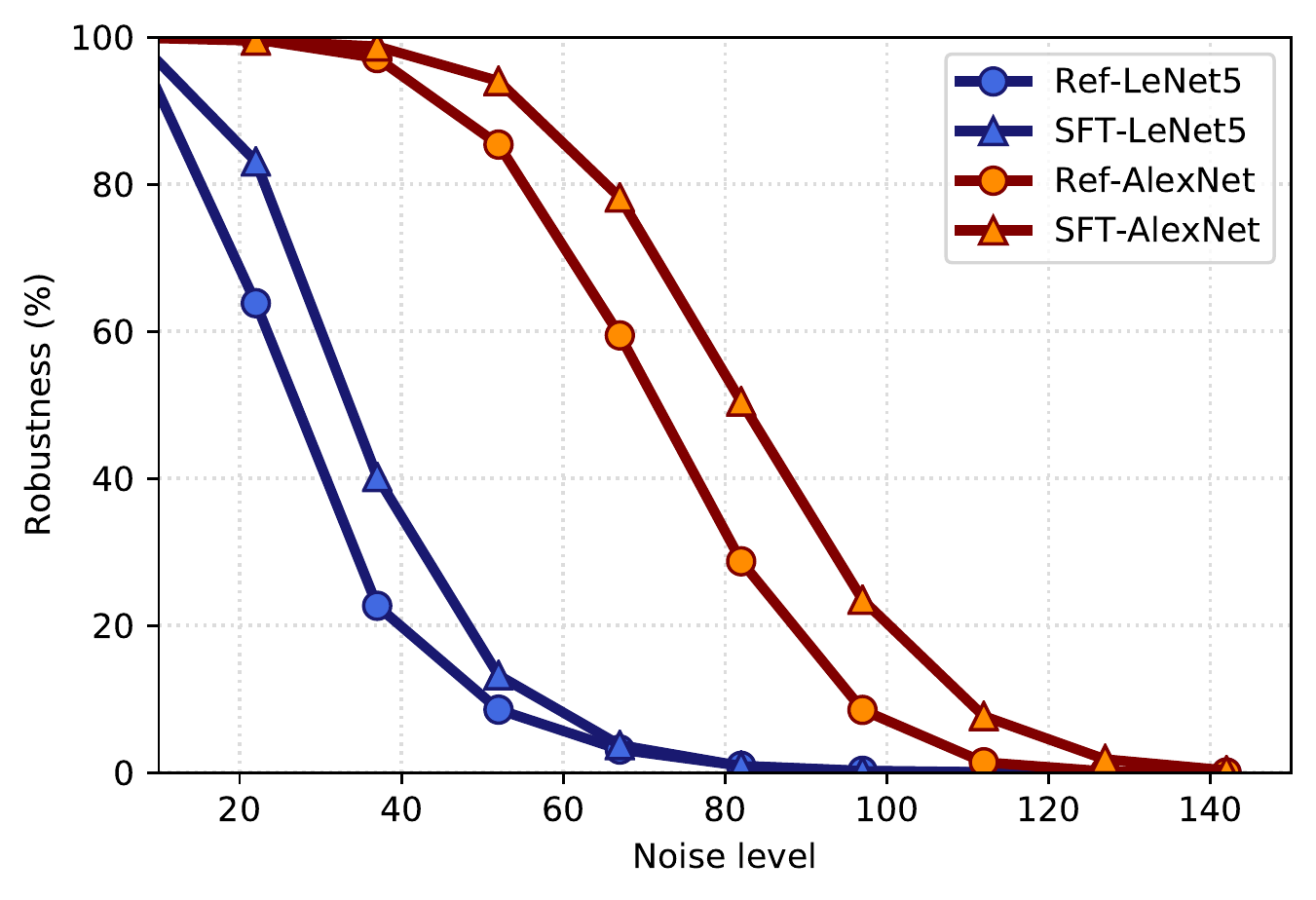}
		
	}\subfloat[RLines]{\includegraphics[width=0.25\columnwidth]{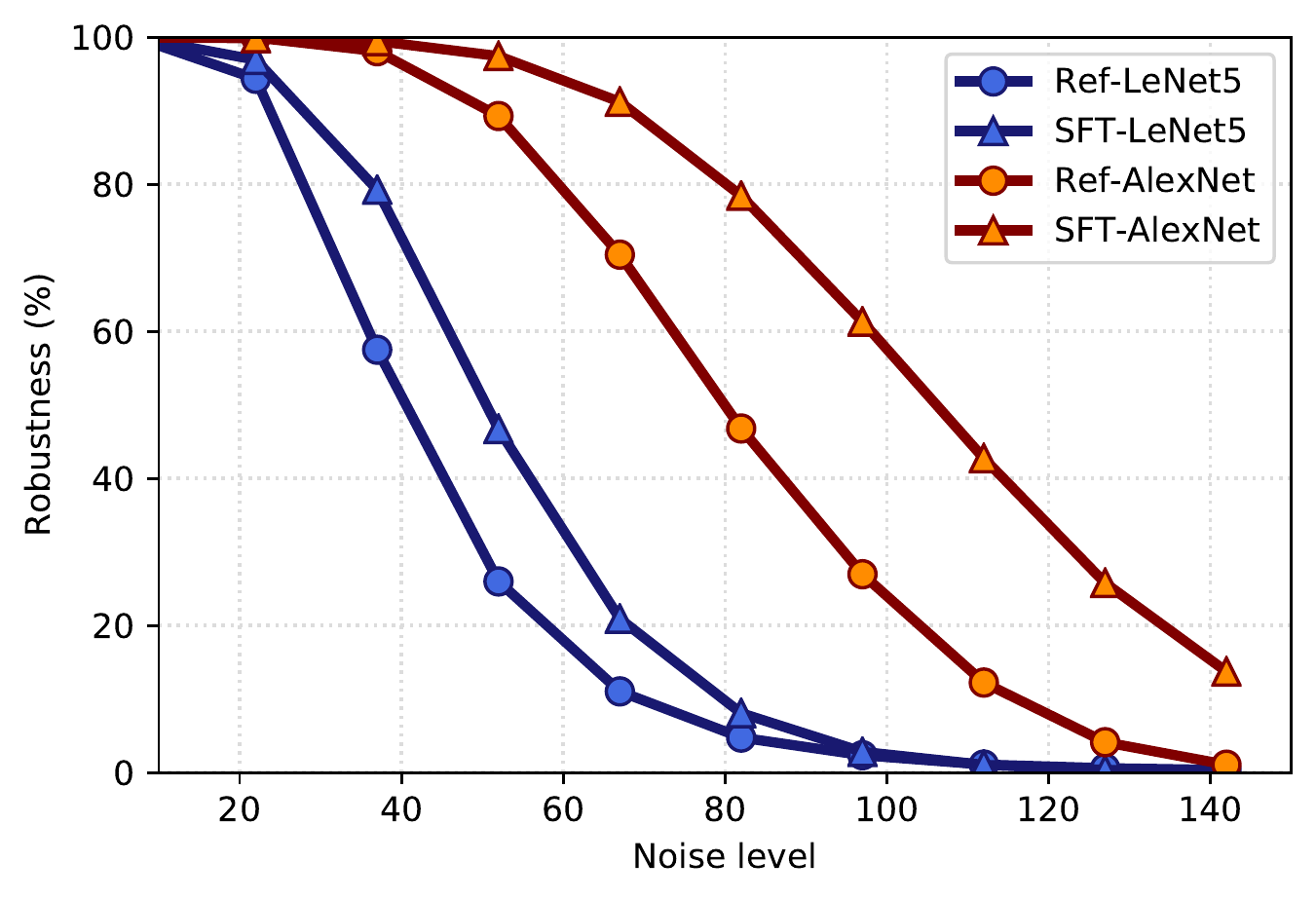}
		
	}\caption{\label{fig:sft:multi-noise:loc}Comparing the effect of different methods of generating contextual noise perturbation on the localization accuracy. From left to right: (a) Grating: radial grating with random centers, (b) MoG: Mixture of Gaussians, (c) Squares: squares with random intensity values, and (d) RLines: short lines with random centers and orientation. The vertical axis represent the localization robustness metric, and the horizontal axis represent the maximum pixel intensity the noise adds to the background.}
	
\end{figure}

These noise methods have chosen such that they cover a variety of shape patterns from small scale to large scale with different line structures and curvatures. We would like to measure the sensitivity of the reference and fine-tuned networks on the samples perturbed with the background noise generated by these methods. Similar to the experiment with the random uniform noise, we report the robustness results to these four noise methods on the LeNet-5 and AlexNet networks for the classification and localization evaluation metrics in Fig. \ref{fig:sft:multi-noise:cls} and Fig. \ref{fig:sft:multi-noise:loc} respectively.

The results reveal that the generalization against contextual noise achieved by the proposed fine-tuning method is persistent across all of the four noise sources for classification and localization. The robustness for Grating and Squares is less than for RLines due to the larger scale of noise patterns. Similar to the uniform noise patterns, RLines have small scale random elements. Both Grating and Squares show consistent robustness gain once the selective fine-tuning is used. MoG is the only method that benefits from smooth and continuous shape patterns. The proposed method still provides slight robustness gain in comparison with the reference networks. Though, the gap is small for AlexNet, we observe improvement for LeNet-5. 

\begin{figure}
	\centering
	\includegraphics[width=0.10\columnwidth]{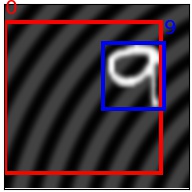}
	\includegraphics[width=0.10\columnwidth]{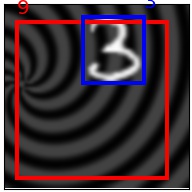}
	\includegraphics[width=0.10\columnwidth]{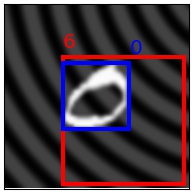}
	\includegraphics[width=0.10\columnwidth]{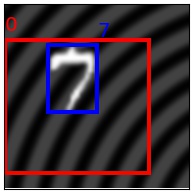}
	\includegraphics[width=0.10\columnwidth]{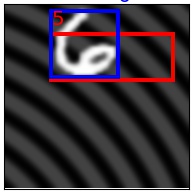}
	\includegraphics[width=0.10\columnwidth]{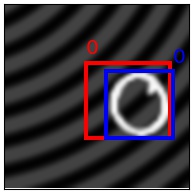}
	\includegraphics[width=0.10\columnwidth]{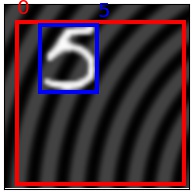}
	\includegraphics[width=0.10\columnwidth]{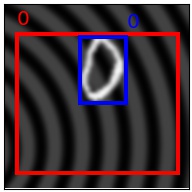}\\
	\vspace{0.005\columnwidth}
	\includegraphics[width=0.10\columnwidth]{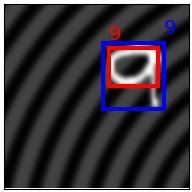}
	\includegraphics[width=0.10\columnwidth]{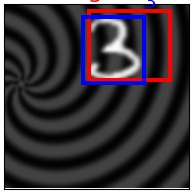}
	\includegraphics[width=0.10\columnwidth]{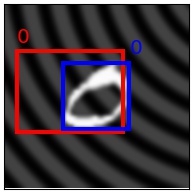}
	\includegraphics[width=0.10\columnwidth]{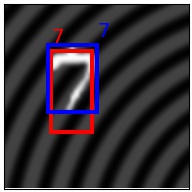}
	\includegraphics[width=0.10\columnwidth]{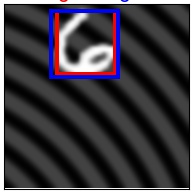}
	\includegraphics[width=0.10\columnwidth]{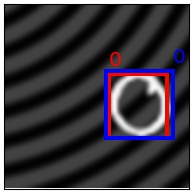}
	\includegraphics[width=0.10\columnwidth]{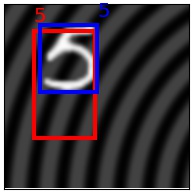}
	\includegraphics[width=0.10\columnwidth]{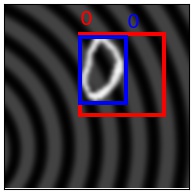}\\		
	{\scriptsize (a) Grating}
	\vspace{0.01\columnwidth}
	
	\includegraphics[width=0.10\columnwidth]{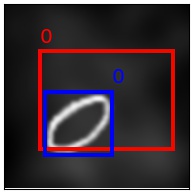}
	\includegraphics[width=0.10\columnwidth]{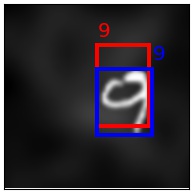}
	\includegraphics[width=0.10\columnwidth]{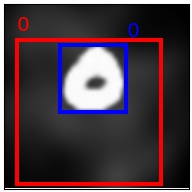}
	\includegraphics[width=0.10\columnwidth]{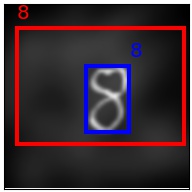}
	\includegraphics[width=0.10\columnwidth]{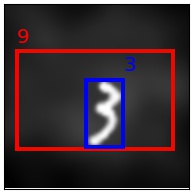}
	\includegraphics[width=0.10\columnwidth]{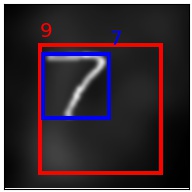}
	\includegraphics[width=0.10\columnwidth]{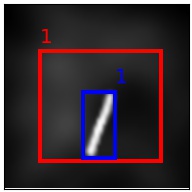}
	\includegraphics[width=0.10\columnwidth]{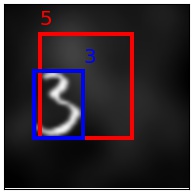}\\
	\vspace{0.005\columnwidth}
	\includegraphics[width=0.10\columnwidth]{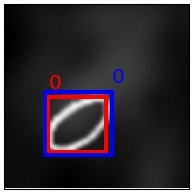}
	\includegraphics[width=0.10\columnwidth]{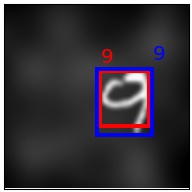}
	\includegraphics[width=0.10\columnwidth]{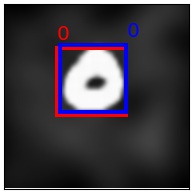}
	\includegraphics[width=0.10\columnwidth]{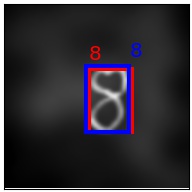}
	\includegraphics[width=0.10\columnwidth]{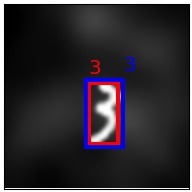}
	\includegraphics[width=0.10\columnwidth]{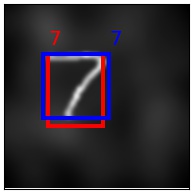}
	\includegraphics[width=0.10\columnwidth]{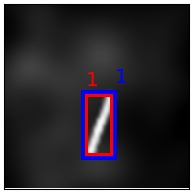}
	\includegraphics[width=0.10\columnwidth]{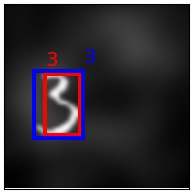}\\		
	{\scriptsize (b) MoG}
	\vspace{0.01\columnwidth}
	
	\includegraphics[width=0.10\columnwidth]{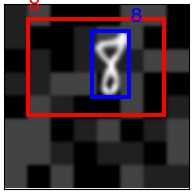}
	\includegraphics[width=0.10\columnwidth]{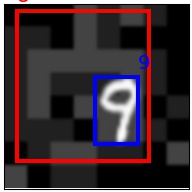}
	\includegraphics[width=0.10\columnwidth]{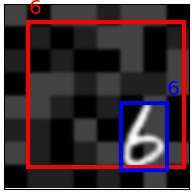}
	\includegraphics[width=0.10\columnwidth]{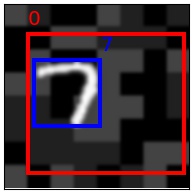}
	\includegraphics[width=0.10\columnwidth]{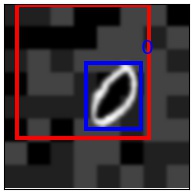}
	\includegraphics[width=0.10\columnwidth]{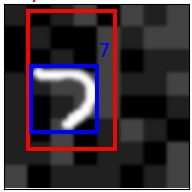}
	\includegraphics[width=0.10\columnwidth]{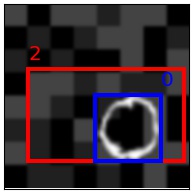}
	\includegraphics[width=0.10\columnwidth]{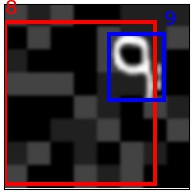}\\
	\vspace{0.005\columnwidth}
	\includegraphics[width=0.10\columnwidth]{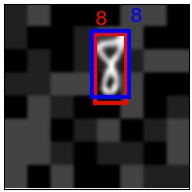}
	\includegraphics[width=0.10\columnwidth]{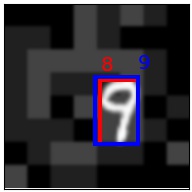}
	\includegraphics[width=0.10\columnwidth]{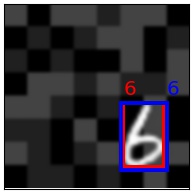}
	\includegraphics[width=0.10\columnwidth]{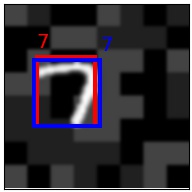}
	\includegraphics[width=0.10\columnwidth]{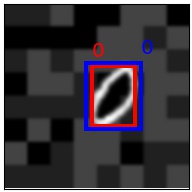}
	\includegraphics[width=0.10\columnwidth]{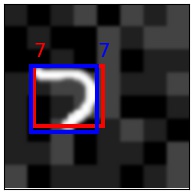}
	\includegraphics[width=0.10\columnwidth]{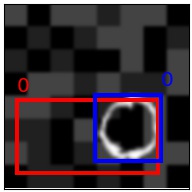}
	\includegraphics[width=0.10\columnwidth]{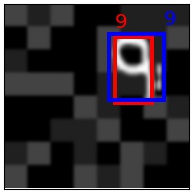}\\		
	{\scriptsize (d) Squares}
	\vspace{0.01\columnwidth}		
	
	\includegraphics[width=0.10\columnwidth]{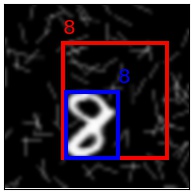}
	\includegraphics[width=0.10\columnwidth]{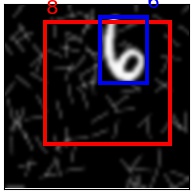}
	\includegraphics[width=0.10\columnwidth]{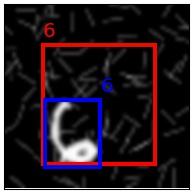}
	\includegraphics[width=0.10\columnwidth]{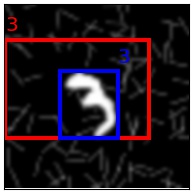}
	\includegraphics[width=0.10\columnwidth]{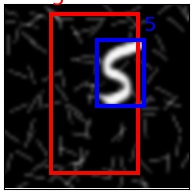}
	\includegraphics[width=0.10\columnwidth]{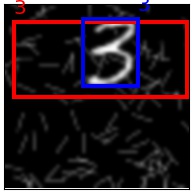}
	\includegraphics[width=0.10\columnwidth]{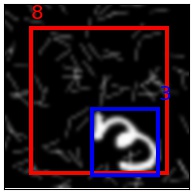}
	\includegraphics[width=0.10\columnwidth]{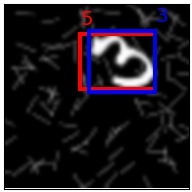}\\
	\vspace{0.005\columnwidth}
	\includegraphics[width=0.10\columnwidth]{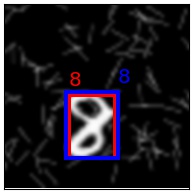}
	\includegraphics[width=0.10\columnwidth]{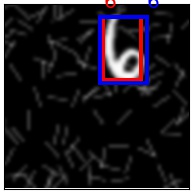}
	\includegraphics[width=0.10\columnwidth]{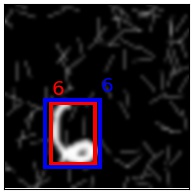}
	\includegraphics[width=0.10\columnwidth]{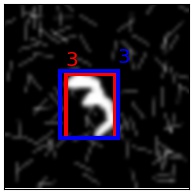}
	\includegraphics[width=0.10\columnwidth]{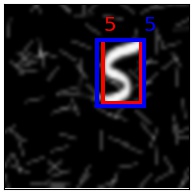}
	\includegraphics[width=0.10\columnwidth]{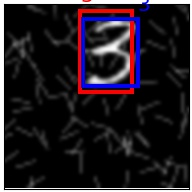}
	\includegraphics[width=0.10\columnwidth]{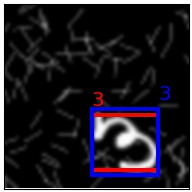}
	\includegraphics[width=0.10\columnwidth]{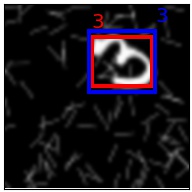}\\		
	{\scriptsize (c) RLines}
	\vspace{0.01\columnwidth}
	
	\caption{\label{fig:sft:multi-noise:alexnet:pred}Comparison of the label and bounding box predictions of the AlexNet reference and fine-tuned networks once the background regions is perturbed with four different types of noise methods. In each section, the top and bottom rows represent predictions from the reference and selective fine-tuned networks. The ground truth and predicted bounding boxes are illustrated with blue and red boxes respectively. The ground truth and predicted labels are shown at the top-left and top-right of their corresponding box respectively.}
	
\end{figure}

The qualitative results for this experimental evaluation setup is illustrated in Fig. \ref{fig:sft:multi-noise:lenet5:pred} and Fig. \ref{fig:sft:multi-noise:alexnet:pred} for LeNet-5 and AlexNet respectively. They show the predicted bounding boxes of the reference and the proposed fine-tuned networks on the perturbed input samples using the four different noise methods. The results support the hypothesis that the TD selective gating method is capable of focusing the learning capacity of the network on the important aspects so then the prediction performance is less affected by the contextual perturbations. The qualitative results illustrates the cases in which the reference network fails to predict the class labels and bounding boxes accurately due to the background noise disturbance. On the other hand, the selective fine-tuned counterpart provides a more robust representation and it maintains prediction performance despite significant background noise patterns. This generalizes across all four different noise methods for both LeNet-5 and AlexNet networks.

%% file: 05-conclusion.tex
Attention helps humans to learn in distracting and interfering situations.
The selective nature of attentional processes are very well established
in human vision studies. TD approaches are commonly used in the literature to explain the underlying internal believes of a learned representation in neural networks. Following the same research direction, we further investigate the role of the TD approach to focus the learning capacity of a network on the best and most reliable interpretation of given data samples.
We propose a selective learning
method for neural networks that has TD attentive mechanisms. We define
an iterative feedforward pass using the modulation of the first feedforward
pass with the TD gating activities. We experimentally test the impact
of the background context when the network is trained with the proposed
method. The evaluation results on a modified MNIST dataset indicate
that the selection mechanism indeed constrains the learning
capacity of the network on relevant aspects of the visual representation for target semantic abstractions.
Over time, the network parameters converges to the state that has
reduced contextual interference and improved robustness against distortions such as additive noise perturbations of the background regions. The qualitative and quantitative results support the role of the selective fine-tuning using the iterative feedforward pass and the augmented loss function.

%% file: egpaper_for_review.bbl
\begin{thebibliography}{10}\itemsep=-1pt

\bibitem{akhtar2018threat}
N.~Akhtar and A.~Mian.
\newblock Threat of adversarial attacks on deep learning in computer vision: A
  survey.
\newblock {\em arXiv preprint arXiv:1801.00553}, 2018.

\bibitem{alistarh2018convergence}
D.~Alistarh, T.~Hoefler, M.~Johansson, N.~Konstantinov, S.~Khirirat, and
  C.~Renggli.
\newblock The convergence of sparsified gradient methods.
\newblock In {\em Advances in Neural Information Processing Systems}, pages
  5975--5985, 2018.

\bibitem{biparva2017stnet}
M.~Biparva and J.~K. Tsotsos.
\newblock Stnet: selective tuning of convolutional networks for object
  localization.
\newblock In {\em ICCV Workshops}, pages 2715--2723, 2017.

\bibitem{chen2016deeplab}
L.-C. Chen, G.~Papandreou, I.~Kokkinos, K.~Murphy, and A.~L. Yuille.
\newblock {Deeplab: Semantic image segmentation with deep convolutional nets,
  atrous convolution, and fully connected crfs}.
\newblock {\em arXiv preprint arXiv:1606.00915}, 2016.

\bibitem{chen2018deeplab}
L.-C. Chen, G.~Papandreou, I.~Kokkinos, K.~Murphy, and A.~L. Yuille.
\newblock Deeplab: Semantic image segmentation with deep convolutional nets,
  atrous convolution, and fully connected crfs.
\newblock {\em IEEE Transactions on Pattern Analysis and Machine Intelligence},
  40(4):834--848, 2018.

\bibitem{dodge2017can}
S.~Dodge and L.~Karam.
\newblock Can the early human visual system compete with deep neural networks?
\newblock In {\em Proceedings of the IEEE International Conference on Computer
  Vision}, pages 2798--2804, 2017.

\bibitem{gilbert2013top}
C.~D. Gilbert and W.~Li.
\newblock Top-down influences on visual processing.
\newblock {\em Nature Reviews Neuroscience}, 14(5):350--363, 2013.

\bibitem{he2017mask}
K.~He, G.~Gkioxari, P.~Doll{\'a}r, and R.~Girshick.
\newblock Mask r-cnn.
\newblock In {\em Proceedings of the IEEE international conference on computer
  vision}, pages 2961--2969, 2017.

\bibitem{he2016deep}
K.~He, X.~Zhang, S.~Ren, and J.~Sun.
\newblock Deep residual learning for image recognition.
\newblock In {\em IEEE Conference on Computer Vision and Pattern Recognition},
  pages 770--778, 2016.

\bibitem{huang2017densely}
G.~Huang, Z.~Liu, L.~Van Der~Maaten, and K.~Q. Weinberger.
\newblock Densely connected convolutional networks.
\newblock In {\em Proceedings of the IEEE conference on computer vision and
  pattern recognition}, pages 4700--4708, 2017.

\bibitem{hupe1998cortical}
J.~Hup{\'e}, A.~James, B.~Payne, S.~Lomber, P.~Girard, and J.~Bullier.
\newblock Cortical feedback improves discrimination between figure and
  background by v1, v2 and v3 neurons.
\newblock {\em Nature}, 394(6695):784, 1998.

\bibitem{islam2017gated}
M.~A. Islam, M.~Rochan, N.~D. Bruce, and Y.~Wang.
\newblock Gated feedback refinement network for dense image labeling.
\newblock In {\em 2017 IEEE Conference on Computer Vision and Pattern
  Recognition (CVPR)}, pages 4877--4885. IEEE, 2017.

\bibitem{krizhevsky2012imagenet}
A.~Krizhevsky, I.~Sutskever, and G.~E. Hinton.
\newblock Imagenet classification with deep convolutional neural networks.
\newblock In {\em Advances in Neural Information Processing Systems}, pages
  1097--1105, 2012.

\bibitem{kurakin2016adversarial}
A.~Kurakin, I.~Goodfellow, and S.~Bengio.
\newblock Adversarial examples in the physical world.
\newblock {\em arXiv preprint arXiv:1607.02533}, 2016.

\bibitem{Lecun1998}
Y.~Lecun, L.~Bottou, Y.~Bengio, and P.~Haffner.
\newblock Gradient-based learning applied to document recognition.
\newblock {\em Proceedings of the IEEE}, 86(11):2278--2324, Nov 1998.

\bibitem{liu2016ssd}
W.~Liu, D.~Anguelov, D.~Erhan, C.~Szegedy, S.~Reed, C.-Y. Fu, and A.~C. Berg.
\newblock Ssd: Single shot multibox detector.
\newblock In {\em European Conference on Computer Vision}, pages 21--37.
  Springer, 2016.

\bibitem{long2015fully}
J.~Long, E.~Shelhamer, and T.~Darrell.
\newblock Fully convolutional networks for semantic segmentation.
\newblock In {\em IEEE Conference on Computer Vision and Pattern Recognition},
  pages 3431--3440, 2015.

\bibitem{nguyen2015deep}
A.~Nguyen, J.~Yosinski, and J.~Clune.
\newblock Deep neural networks are easily fooled: High confidence predictions
  for unrecognizable images.
\newblock In {\em IEEE Conference on Computer Vision and Pattern Recognition},
  pages 427--436, 2015.

\bibitem{paszke2017automatic}
A.~Paszke, S.~Gross, S.~Chintala, G.~Chanan, E.~Yang, Z.~DeVito, Z.~Lin,
  A.~Desmaison, L.~Antiga, and A.~Lerer.
\newblock Automatic differentiation in pytorch.
\newblock In {\em NIPS 2017 Autodiff Workshop: The Future of Gradient-based
  Machine Learning Software and Techniques}, 2017.

\bibitem{redmon2016yolo9000}
J.~Redmon and A.~Farhadi.
\newblock {YOLO9000: Better, Faster, Stronger}.
\newblock {\em arXiv preprint arXiv:1612.08242}, 2016.

\bibitem{ren2015faster}
S.~Ren, K.~He, R.~Girshick, and J.~Sun.
\newblock Faster r-cnn: Towards real-time object detection with region proposal
  networks.
\newblock In {\em Advances in Neural Information Processing Systems}, pages
  91--99, 2015.

\bibitem{rosenfeld2018elephant}
A.~Rosenfeld, R.~Zemel, and J.~K. Tsotsos.
\newblock The elephant in the room.
\newblock {\em arXiv preprint arXiv:1808.03305}, 2018.

\bibitem{simonyan2013deep}
K.~Simonyan, A.~Vedaldi, and A.~Zisserman.
\newblock Deep inside convolutional networks: Visualising image classification
  models and saliency maps.
\newblock {\em arXiv preprint arXiv:1312.6034}, 2013.

\bibitem{sun2017meprop}
X.~Sun, X.~Ren, S.~Ma, and H.~Wang.
\newblock meprop: Sparsified back propagation for accelerated deep learning
  with reduced overfitting.
\newblock {\em arXiv preprint arXiv:1706.06197}, 2017.

\bibitem{szegedy2014going}
C.~Szegedy, W.~Liu, Y.~Jia, P.~Sermanet, S.~Reed, D.~Anguelov, D.~Erhan,
  V.~Vanhoucke, and A.~Rabinovich.
\newblock Going deeper with convolutions.
\newblock In {\em IEEE Conference on Computer Vision and Pattern Recognition},
  pages 1--9, 2015.

\bibitem{tsotsos2011computational}
J.~K. Tsotsos.
\newblock {\em A computational perspective on visual attention}.
\newblock MIT Press, 2011.

\bibitem{tsotsos1995SelTun}
J.~K. Tsotsos, S.~M. Culhane, W.~Y.~K. Wai, Y.~Lai, N.~Davis, and F.~Nuflo.
\newblock Modeling visual attention via selective tuning.
\newblock {\em Artificial Intelligence}, 78(1--2):507--545, 1995.
\newblock Special Volume on Computer Vision.

\bibitem{tsotsos2008different}
J.~K. Tsotsos, A.~J. Rodr{\'\i}guez-S{\'a}nchez, A.~L. Rothenstein, and
  E.~Simine.
\newblock The different stages of visual recognition need different attentional
  binding strategies.
\newblock {\em Brain research}, 1225:119--132, 2008.

\bibitem{varga2017gradient}
D.~Varga, A.~Csisz{\'a}rik, and Z.~Zombori.
\newblock Gradient regularization improves accuracy of discriminative models.
\newblock {\em arXiv preprint arXiv:1712.09936}, 2017.

\end{thebibliography}
